\documentclass[sigconf, nonacm]{acmart}
 
\AtBeginDocument{%
  \providecommand\BibTeX{{%
    \normalfont B\kern-0.5em{\scshape i\kern-0.25em b}\kern-0.8em\TeX}}}

\usepackage{multirow}
\usepackage{makecell}
\usepackage{colortbl}
\usepackage[normalem]{ulem}
\usepackage{xcolor}
\usepackage{amsfonts}
\usepackage{graphicx}
\usepackage{subcaption}
\usepackage{lipsum} %
\usepackage{float} %
\usepackage{nicematrix}
\definecolor{LightGreen}{RGB}{2, 125, 86} %

\useunder{\uline}{\ul}{}

\begin{document}

\title{MiniGPT-3D: Efficiently Aligning 3D Point Clouds with 
  Large Language Models using 2D Priors}

\author{Yuan Tang$^{1}$~\quad
Xu Han$^{1}$ \quad
Xianzhi Li$^{1 \dagger}$~\quad
Qiao Yu$^{1}$~\quad
Yixue Hao$^{1}$~\quad
Long Hu$^{1}$~\quad
Min Chen$^{2}$}

\affiliation{
  \institution{$^{1}$ Huazhong University of Science and Technology \quad 
$^{2}$  South China University of Technology \quad $^{\dagger}$ Corresponding author}
  \city{}
  \country{}}

\affiliation{
  \institution{ \{yuan\_tang,  xhanxu, xzli, qiaoyu\_epic, yixuehao, hulong\}@hust.edu.cn \quad minchen@ieee.org}
  \city{}
  \country{}}

\begin{abstract}
Large 2D vision-language models (2D-LLMs)  have gained significant attention by bridging Large Language Models (LLMs) with images using a simple projector.
Inspired by their success, large 3D point cloud-language models (3D-LLMs) also  integrate point clouds into LLMs.
However, directly aligning point clouds with LLM requires expensive training costs, typically in hundreds of GPU-hours on A100, which hinders the development of 3D-LLMs.
In this paper, we introduce \textbf{MiniGPT-3D}, an efficient and powerful 3D-LLM that achieves multiple  \textbf{SOTA} results while training for only  \textbf{27 hours on one  RTX 3090}.
Specifically, we propose to align 3D point clouds with LLMs using 2D priors from 2D-LLMs, which can leverage the similarity between 2D and 3D visual information. 
We introduce a novel four-stage training strategy for modality alignment in a cascaded way, and a mixture of query experts module to adaptively aggregate features with high efficiency.
Moreover, we utilize parameter-efficient fine-tuning methods  LoRA and  Norm fine-tuning, resulting in only \textbf{47.8M} learnable parameters, which is up to  260$\times$ fewer than existing methods.
Extensive experiments show that MiniGPT-3D achieves  SOTA on 3D object classification and  captioning tasks, with significantly cheaper training costs. 
Notably, MiniGPT-3D gains an 8.12 increase on GPT-4 evaluation score for the  challenging  object captioning task  compared to ShapeLLM-13B, while the latter costs 160  total GPU-hours on 8 A800. 
We are the first to explore the efficient 3D-LLM, offering new insights to the community.
Code and weights are available at  \href{https://github.com/TangYuan96/MiniGPT-3D}{ \emph{https://github.com/TangYuan96/MiniGPT-3D.} }

\end{abstract}


 
\begin{teaserfigure}
  \includegraphics[width=\textwidth]{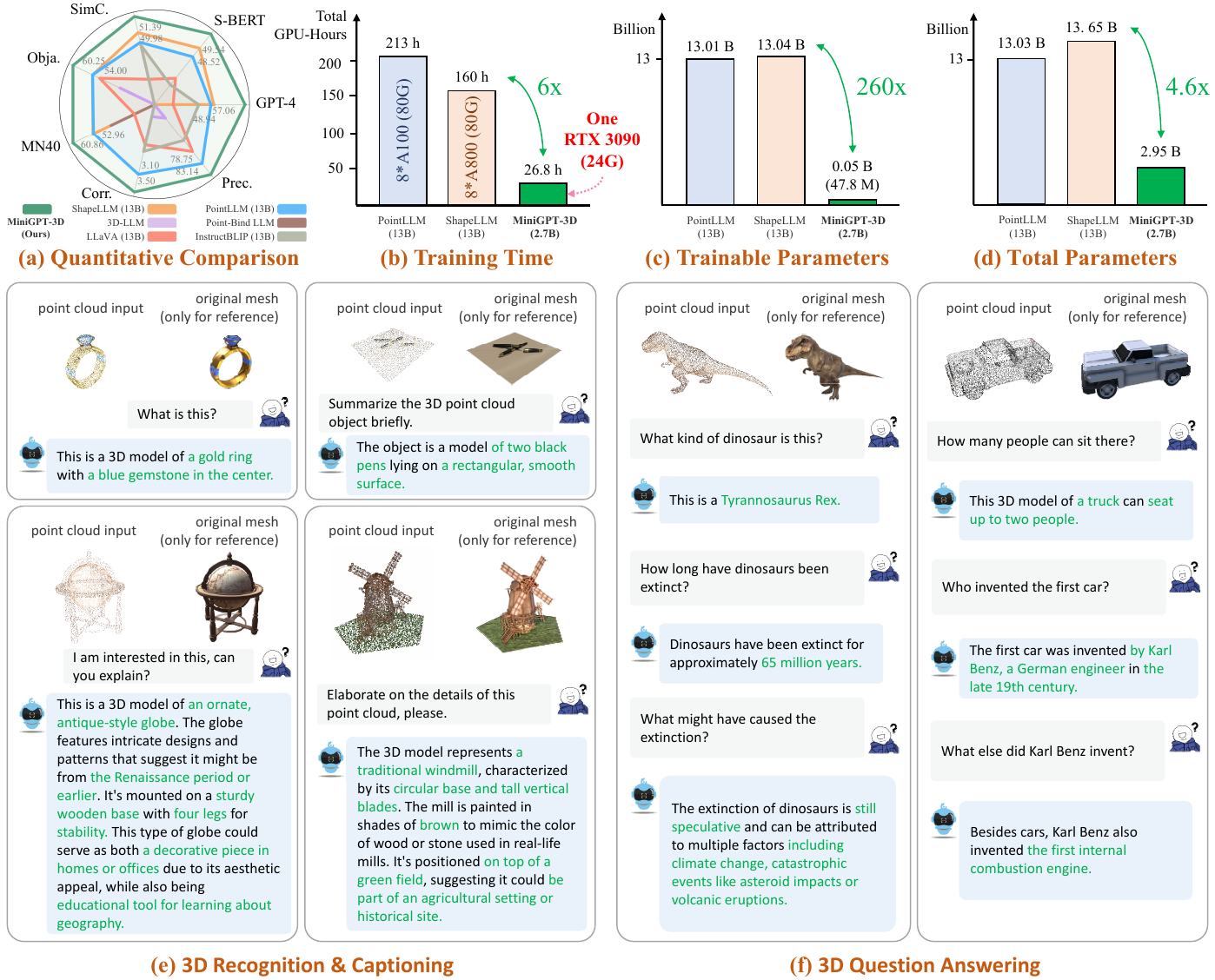}
  \setlength{\abovecaptionskip}{-0.3cm} 
  \caption{Demonstrations of MiniGPT-3D. We present MiniGPT-3D, an efficient and powerful 3D-LLM that aligns 3D point clouds with large language models   using 2D priors from large 2D vision-language models.
This figure demonstrates MiniGPT-3D's superior performance and efficient training compared to existing 3D-LLMs. We also show some prediction examples in 3D recognition, captioning, and question-answering tasks, with the correct and fine-grained answers highlighted in green.}
  \Description{z}
  \label{fig:teaser}
\end{teaserfigure}

\maketitle

\section{Introduction}

\label{sec:intro}
Large Language Models (LLMs) have recently driven advancements in multiple fields~\cite{openai2023gpt, touvron2023llama, thirunavukarasu2023large, fu2024drive}, benefiting from their world knowledge.
Built on LLMs, large 2D vision-language models (2D-LLMs)~\cite{li2023blip, cha2023honeybee, zhu2023minigpt}  can align image features with text through an image feature projector, enabling 2D-LLMs to understand visual content. 
Inspired by 2D-LLMs, large 3D point cloud-language models (3D-LLMs)~\cite{xu2023pointllm, qi2023gpt4point, qi2024shapellm}  aim to incorporate 3D point cloud features into LLMs, equipping LLMs with the ability to perceive and reason in 3D space. 
These 3D-LLMs hold promise for widespread applications in fields like robotics~\cite{singh2023progprompt, wang2024large} and autonomous driving~\cite{fu2024drive, cui2024drive}.
However, 3D-LLMs are expensive to train. 
For example, training PointLLM-13B~\cite{xu2023pointllm} takes 213 total GPU-hours on 8 A100 GPU, making research and applications extremely challenging. 
Here, we aim to find a more efficient way to connect 3D point clouds with LLMs.

\begin{figure}[t]
\centering
  \includegraphics[width=\linewidth]{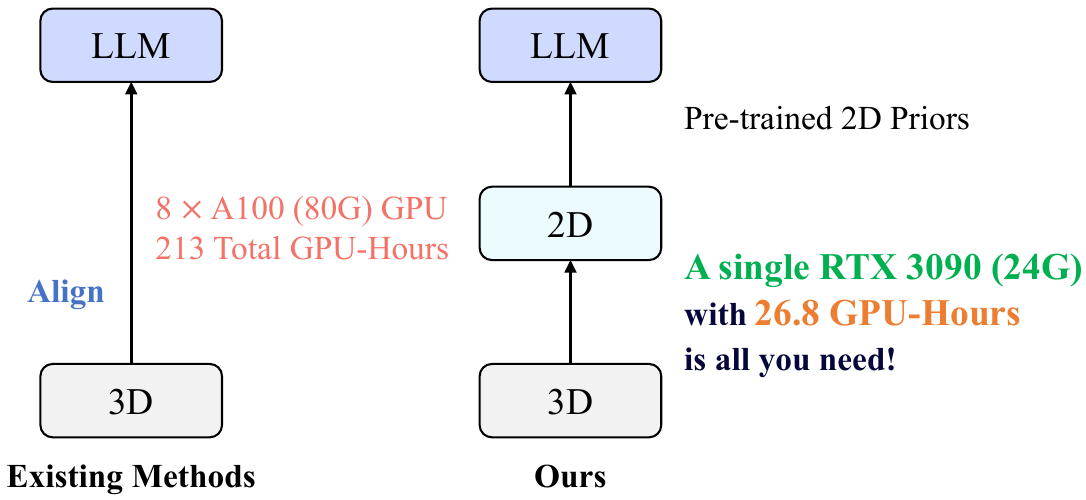}
  \setlength{\abovecaptionskip}{-0.2cm} 
  \caption{Existing methods and  ours   to align 3D with LLMs.}
  \label{fig:motiv}
  \vspace*{-3mm}
\end{figure}

We observe that existing 3D-LLMs directly align point cloud encoders with LLMs. 
Although these encoders can produce somewhat unified features through multimodal pre-training, there is still a significant modality gap between 3D points with LLMs, requiring substantial resources for alignment.
Besides, in contrast to resource-intensive alignment between vision and language, 3D point clouds and 2D images are both visual modalities, which makes it easier to align their representations. 
Thus, we pose a question: \textbf{\textit{Can we use 2D-LLMs as a strong prior to connect LLMs and 3D data, making alignment more efficient?} }
In other words, as shown in Figure~\ref{fig:motiv}, leveraging pre-trained 2D-LLMs directly allows for cutting down the cost of vision-language alignment, leaving only the 2D-3D vision alignment, which is significantly cheaper.

Following this intuition, we propose MiniGPT-3D, an efficient 3D-LLM that connects 3D point clouds and LLMs using 2D-LLMs as priors. Our MiniGPT-3D achieves multiple state-of-the-art (SOTA) results,   requiring only 27 hours of training on a single RTX 3090 GPU.
Specifically,  we propose an efficient four-stage training strategy in a cascaded way, gradually allowing the model to learn unified visual-textual representations.  This process achieves the smooth transfer of priors from 2D-LLM to the 3D space, thus efficiently constructing a bridge from 3D to LLM.
Moreover, we  introduce the Mixture of Query Experts (MQE), which comprises multiple query experts and an expert router, enabling the adaptive aggregation of features from multiple experts with only 0.4M parameters.
MQE dynamically adjusts the cooperation relationship between experts, thereby aggregating 3D features from multiple perspectives into the semantic space of 2D-LLM.
Meanwhile, we employ various parameter-efficient fine-tuning (PEFT)  technologies like LoRA~\cite{hu2021lora} and Norm fine-tuning, and utilize an efficient LLM, further reducing the model's training overhead.

\begin{figure*}[t]
    \centering
  \includegraphics[width=\textwidth]{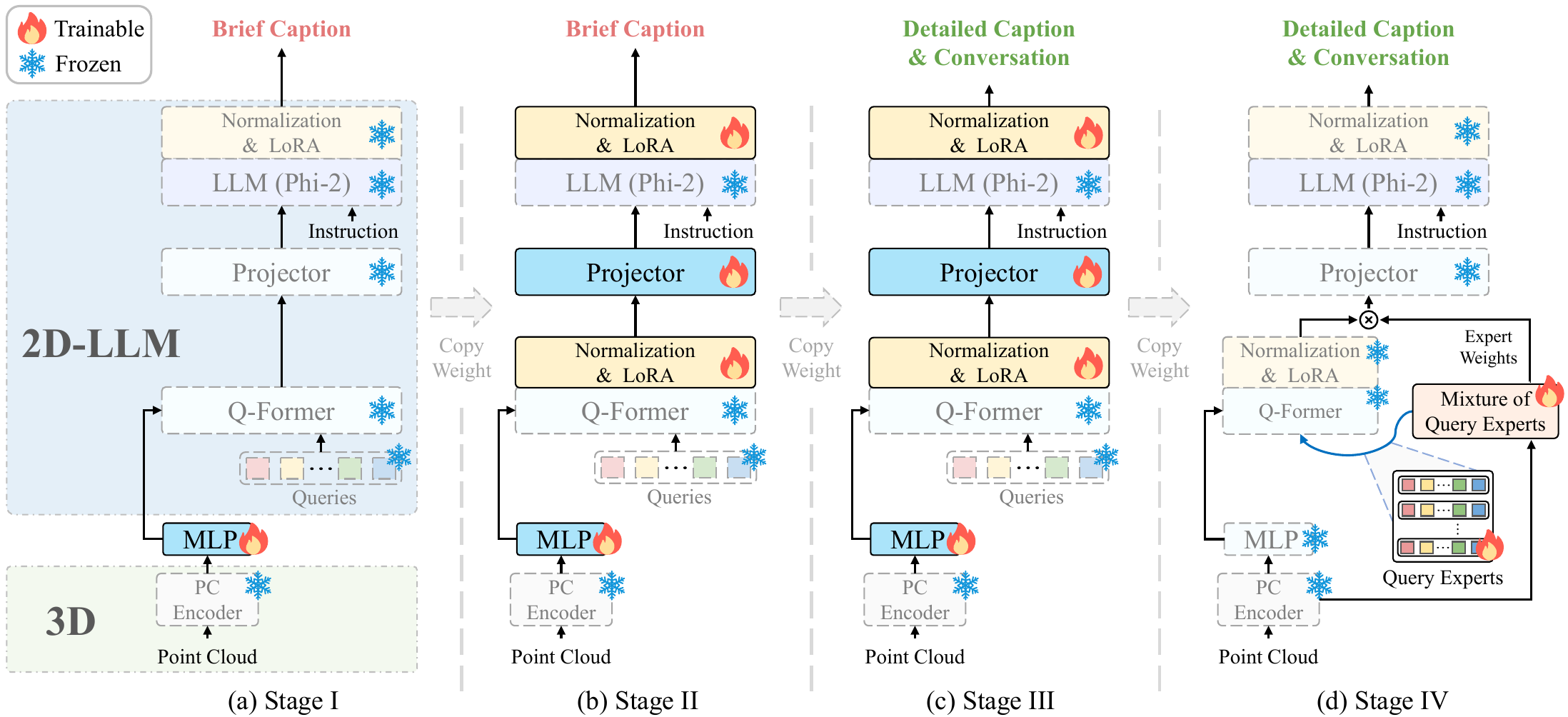}
      \setlength{\abovecaptionskip}{-0.25cm}  
  \caption{
     Training framework and strategy. 
    Our MiniGPT-3D utilizes a four-stage training strategy.
    (a) We  solely train the point cloud projection layer (MLP).
    (b) We  train the modality projector while  fine-tuning the point cloud projection layer, Q-Former, and LLM backbone.
    (c) We further enhance the modules trained in the second stage by leveraging a more challenging task.
    (d) Finally,  we only train the mixture of query experts, while freezing the remaining modules.
    }
  \label{arch}
\end{figure*}

As shown in Figure~\ref{fig:teaser},   MiniGPT-3D achieves new SOTA performance on generative 3D object classification and object captioning tasks. Specifically, compared to the powerful baseline ShapeLLM-13B~\cite{qi2024shapellm}, MiniGPT-3D achieves a 6.77\% increase in classification average accuracy and an 8.12 increase in GPT-4 evaluation score.
Notably, MiniGPT-3D utilizes extremely cheaper training resources (1$\times$ RTX 3090 vs. 8$\times$ A800), with up to 6$\times$ acceleration (26.8h on RTX 3090 vs. 160h on A800). Furthermore, our model has significantly fewer trainable parameters, reduced by up to 260$\times$, with 2.95B model parameters in total, which is decreased by up to 4.6$\times$.

MiniGPT-3D takes the first step in efficient 3D-LLM, we hope that MiniGPT-3D can bring new insights to this community. In summary, our contributions are as follows:

\begin{itemize}
    \item We present MiniGPT-3D, an efficient and powerful 3D-LLM that  aligns 3D points with LLMs  using 2D priors, achieving multiple SOTA with only 26.8h of training on one RTX 3090.  
    \item We propose an efficient four-stage training strategy in a cascaded way, gradually transferring the knowledge from 2D-LLMs to 3D while requiring only 47.8M learnable parameters.
    \item We design the mixture of query experts to aggregate multiple features from different experts with only 0.4M parameters.
    \item Extensive experiments show  the  superior performance of MiniGPT-3D on multiple tasks  while reducing the training time and parameters by up to 6x and 260x, respectively. 
\end{itemize}

\section{Related Work}
\label{sec:rw}

\subsection{Large 2D Vision-Language Models}

    The exceptional instruction-following and generalization capabilities of    LLMs~\cite{touvron2023llama,wei2022chain,ye2023mplug,yang2023baichuan} have been integrated into vision, leading to the emergence of large 2D vision-language models (2D-LLMs).
    Early works such as Flamingo~\cite{alayrac2022flamingo} and BLIP-2~\cite{li2023blip} successfully use projectors to align vision information to LLMs.
    More recently, most works mainly focus on improving model capabilities through expanding the instruction-tuning dataset~\cite{liu2023aligning, zhang2023enhanced, chen2023sharegpt4v}, increasing resolution of image~\cite{liu2023improved, bai2023qwen}, enhancing image encoders~\cite{zhang2023internlm, chen2023internvl}.
    Meanwhile, some methods~\cite{chu2023mobilevlm, chu2024mobilevlm, zhang2024tinyllama, yuan2023tinygpt} have also begun to explore efficient 2D-LLM.
    Models like TinyLlama~\cite{zhang2024tinyllama} and TinyGPT-V~\cite{yuan2023tinygpt} use Phi-2~\cite{Phi_2}, an efficient LLM, to achieve easily deployable 2D-LLMs.
    Among them,  TinyGPT-V leverages LoRA~\cite{hu2021lora} technology and pre-trained  modules to achieve extremely efficient fine-tuning.
    However, TinyGPT-V can only handle 2D images, efficient 3D-LLM remains unexplored, and we aim to fill this gap.

\vspace*{-2.3mm}

\subsection{Large 3D  Point Cloud-Language Models}

    Large 3D point cloud-language models (3D-LLMs)  introduce LLM into the  point cloud modality~\cite{hong20243d,xu2023pointllm,qi2024shapellm, ji2023jm3d, qi2023gpt4point, panagopoulou2023x,   chen2023ll3da,liu2024uni3d, yang2023lidar}.
    Early attempt~\cite{hong20243d} renders 3D objects into 2D multi-view images, then utilizes 2D-LLM to understand 3D.
    However,  the absence of  direct perception of raw point cloud data limits its comprehension of 3D geometry. 
    To address this issue, recent works~\cite{ji2023jm3d, qi2023gpt4point, panagopoulou2023x,  chen2023ll3da} propose to discard the ``rendering'' and encode point cloud directly, followed by modal alignment to fixed LLMs via trainable projectors.
    PointLLM~\cite{xu2023pointllm} and ShapeLLM~\cite{qi2024shapellm} show that  models can be enhanced after fully fine-tuning.
    However, the training of 3D-LLMs is expensive.
    For instance, PointLLM-13B requires training on 8 A100 GPUs for up to 216  total GPU-hours.
    We observe that with 2D-LLM as visual prior, we can not only bypass the ``point cloud rendering'', but also make this hierarchical alignment extremely efficient. 
    Therefore, we propose MiniGPT-3D, different from existing 3D-LLMs which aligns 3D points directly to LLMs, our MiniGPT-3D leverages the powerful priors from 2D-LLM as a linkage between LLM and 3D points, using only a  RTX 3090  to train  for 27 hours.

\vspace*{-2mm}

\subsection{Mixture of Experts}

    Mixture of Experts (MoE)~\cite{jacobs1991adaptive, jordan1994hierarchical} is an ensemble learning technique that adaptively activates selected modules, referred to as experts, based on input.
    MoE is widely used in various fields~\cite{shazeer2017outrageously, lepikhin2020gshard, kudugunta2021beyond, fedus2022switch, riquelme2021scaling}.
    ~\citeauthor{shazeer2017outrageously} introduce MoE into  NLP for the first time, where each intervening layer between  LSTM layers serves as an expert.
    Gshard~\cite{lepikhin2020gshard} further  expands the  MoE to Transformer~\cite{vaswani2017attention}, treating each Feed-Forward Neural Network (FNN) as an expert.
    Recently, with the emergence of LoRA, several works~\cite{gao2024higher,zadouri2023pushing,dou2023loramoe}  design   FFN's LoRA network  as an expert  to efficiently fine-tune LLM.
    Moreover, OneLLM~\cite{han2023onellm} introduces MoE to the learned projector of 2D-LLM, with each projector serving as an expert.
    In our work, we integrate the MoE concept into the queries  of  Q-Former~\cite{li2023blip}, treating each set of queries as an  expert. These experts   adaptively aggregate  point cloud features across diverse extraction perspectives.

\vspace*{-2mm}

\section{Method}
\label{sec:method}
    In this section, we first introduce the architecture of  MiniGPT-3D  (Sec.~\ref{Model_Architecture}), 
    and then present our four-stage training strategy (Sec.~\ref{Training_Stages}), 
    and  finally elucidate the training loss for  MiniGPT-3D (Sec.~\ref{Training_Loss}).

\subsection{Model Architecture}
    \label{Model_Architecture}

    Figure~\ref{arch} depicts the architecture of MiniGPT-3D, which consists of the six main components: a point cloud encoder,  a point cloud projection layer (MLP), a Q-Former,  a mixture of query expert (MQE), a modality projector, and a large language model.
    
    The MiniGPT-3D framework introduces a two-step projection process,  transforming the point cloud from 3D to 2D and then to 1D.
    Specifically, the point cloud is passed to the point cloud encoder to extract 3D features.
    Subsequently, features are then projected into a 2D semantic space using the point cloud projection layer.
    Finally, leveraging the 2D-LLM modules  including the Q-Former, modality projector, Norm of LLM, and LoRA of LLM,   features in 2D-LLM space are transduced  into the 1D-text space of LLM, enabling efficient alignment between 3D and LLM.
    Additionally, MQE enhances MiniGPT-3D's comprehensive and accurate perception of 3D objects.
    Details are presented in the following sections.
    
\subsubsection{\textbf{3D Features  to 2D}} During this process, the point cloud is encoded into 3D features and subsequently projected into the 2D semantic space of the 2D-LLM.

    \paragraph{\textbf{Point Cloud Encoder}} The input point cloud  is encoded into 3D features by the point cloud encoder $f_{pc}$. 
    Specifically, the point cloud $P\in\mathbb{R}^{n\times d}$ is input to $f_{pc}$ , where ${n}$ is the number of points  and ${d}$ denotes the feature dimension of each point.
    Then,   $f_{pc}$ outputs a point feature sequence $X\in\mathbb{R}^{m\times b}$, comprising $m$ features, each with a dimension of $b$.
    In our experiments, we employ  the Point-BERT~\cite{yu2022point} model, pre-trained on ULIP-2~\cite{xue2023ulip} using the Objaverse~\cite{deitke2023objaverse} dataset, as the point cloud encoder. 
    To maintain pre-training knowledge, we freeze the encoder's parameters on all training stages.

    \paragraph{\textbf{Point Cloud Projection Layer}} The point cloud projection layer $f_{MLP}$ is an MLP with two linear layers, which embeds point features $X$  into the semantic space of the pre-trained 2D Q-Former~\cite{li2023blip}, aligning their dimensions. 
    Concisely, $ Y = f_{MLP}(X)$, where $Y \in\mathbb{R}^{m\times b^{\prime}}$ and $b^{\prime}$ is the    hidden space dimension of Q-Former.

 \begin{figure}[t]
\centering
    \setlength{\abovecaptionskip}{0.1cm} 
  \includegraphics[width=\linewidth]{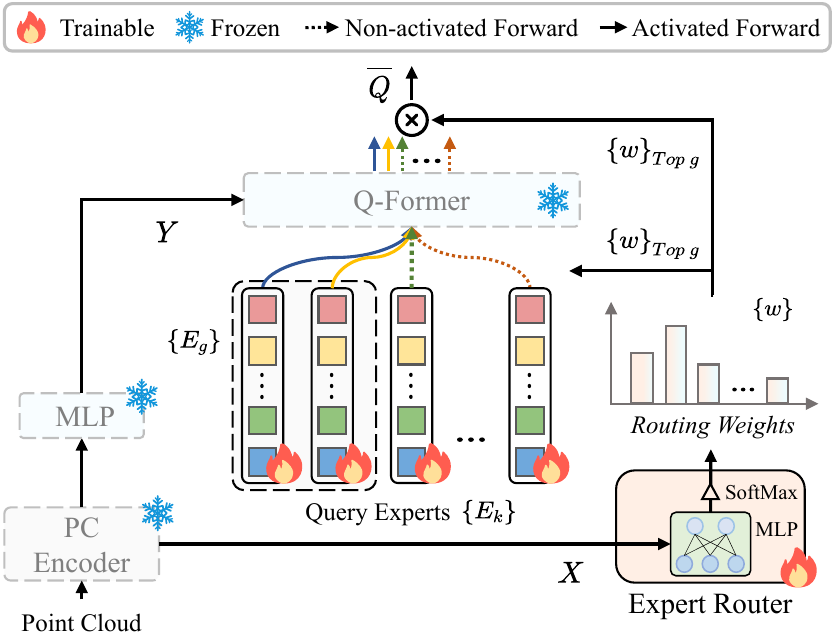}
  \caption{
    The framework of the mixture of query experts.
    First, a point cloud is encoded to  features $X$ and $Y$. 
    Feature $X$  is then passed through  to the expert router, assigning softmax-based weights to experts. 
    The top $g$ experts are selected based on these weights.
    These experts, together with $Y$,  are then fed into the Q-Former, and their outputs are weighted to produce the final point queries $\overline{Q}$.
}
 
  \label{query_expert}
  \vspace*{-2mm}
\end{figure}

\begin{table*}[t]
    \centering
    \setlength\tabcolsep{12pt}
    \setlength{\abovecaptionskip}{0.1cm}  
    \renewcommand{\arraystretch}{0.87}
    
    \caption{Each training stage setups and overhead.}
    \resizebox*{\linewidth}{!}{
    
    \begin{tabular}{lccccrc}
        \toprule
        Training Stages & Dataset Types     & Dataset Scale &  Epochs & Init\_lr \& Min\_lr   &\begin{tabular}[c]{@{}c@{}}Trainable \\ Parameters\end{tabular}& \begin{tabular}[c]{@{}c@{}}Training Time using \\  One RTX 3090 GPU\end{tabular} \\ \midrule
        Stage I         & Brief   Caption                  & 660 k                                                            &   1         & 3e-5, 1e-5   & 1.4 M  & 9.4 h                                                                            \\
        Stage II         & Brief   Caption                  & 660 k                                                          &         1     & 3e-5, 1e-5  & 47.4 M    & 10.9 h                                                                           \\
        Stage III         & Detailed Caption \& Conversation & 70 k                                                           &      3     & 1e-5, 1e-6   & 47.4 M      & 4.9 h                                                                            \\
        Stage  IV         & Detailed Caption \& Conversation & 70 k                                                             &    1      & 5e-6, 1e-6    & 0.4 M     & 1.6 h                                                                            \\ \bottomrule
    \end{tabular}
}
    
    \label{all_stages}
\end{table*}

\subsubsection{\textbf{Features in  2D-LLM space to LLM}} This part transduces  the point cloud representation in the 2D semantic space of   2D-LLM   to the 1D text space of LLM.

    \paragraph{\textbf{Q-Former}} The Q-Former $f_{QF}$, with a decoder-based Transformer structure,  transforms point  features $Y$ into  point queries $\overline{Q}$. 
    This process not only enhances the information extracted from point cloud features  but also reduces  input size for subsequent LLM, accelerating training and inference. 
    Concisely,  $\overline{Q} = f_{QF}(Y,~Q)$, where  $Q \in\mathbb{R}^{o\times b^{\prime}}$, $\overline{Q} \in\mathbb{R}^{o\times b^{\prime}}$. $Q$ is the queries of Q-former and $o$ is  the  number of query. 
    In  experiments, we initialize  Q-Former with   BLIP-2~\cite{li2023blip} pre-trained weights. 
    Given Q-Former's extensive 105M parameters, we employ PEFT technologies to fine-tune its  Query, Key, and Value layers, and normalization layers, thus enhancing adaptability to point clouds while preserving  2D  knowledge.

    \paragraph{\textbf{Mixture of Query Experts}} Inspired by multi-view image rendering for 3D-to-2D projection, we propose the Mixture of Query Experts (MQE) to achieve a similar effect. 
     In the process of MQE, multiple sets of queries (query expert) are used  to transform point features into the semantic space of  2D Q-Former. 
     MQE is the first to introduce dynamic routing of MoE into queries, enabling adaptive activation of more suitable query experts  to capture richer semantic information across diverse point cloud inputs, as shown in Figure~\ref{query_expert}.
     MQE contains $k$ trainable query experts $\{E_k\}$, each is a set of queries  initialized from BLIP-2.
    To integrate multiple query experts into one set of  queries, we use a dynamic routing,  expert router $f_R$, which regulates each expert's contribution. 
    The expert router includes  an MLP and a softmax operation, which  accepts feature $X$ and assigns   softmax-based routing weights to each expert. 
    We employ the sparse routing strategy~\cite{shazeer2017outrageously}, selecting   $g$ experts with the highest weights. 
    Next, the selected query experts $\{E_g\}$  utilize Q-Former to extract high-dimensional semantics $\{\overline{Q_h}\}$ from the feature $Y$. 
    $\{\overline{Q_h}\}$ are  then weighted by the corresponding routing weights to generate the final point queries $\overline{Q}$. 
    The process can be formulated as: 
    %
    \begin{align}
         \overline{Q} &=\sum_{E_q\in\{E_g\}} w_q \cdot  f_{QF}(Y,~E_q), \\
          w_q  &= f_R(X) \left [ q \right ] = \text{Softmax}\left(\text{MLP}(X)\right) \left [ q \right ].
    \end{align}
    To enable query experts to learn knowledge within a stable 3D-LLM semantic context, MQE is only utilized  in the final training stage, by which time other modules have completed training.

     \paragraph{\textbf{Modality Projector}} We use an MLP as the modality projector to  bridge the modality gap between point cloud and text, while  transforming point  queries $\overline{Q} \in \mathbb{R}^{o\times b^{\prime}}$ into point tokens $ T_{pc} \in \mathbb{R}^{o\times c}$, where $c$ denotes  the shared dimension of both point and text tokens.
    
\subsubsection{\textbf{Large Lanuguage Model Backbone}}

    To minimize GPU memory usage during training, we utilize Phi-2~\cite{Phi_2} with 2.7 billion parameters as the large language model backbone of MiniGPT-3D.

    In MiniGPT-3D, the LLM backbone $f_{llm}$  processes a sequence of tokens $T=\left(t_1, t_2, \ldots, t_j\right) \in \mathbb{R}^{j \times c}$, where $j$ is the number of tokens,  including  point tokens and text tokens.
    Leveraging  the  self-attention mechanism,  the LLM backbone  can   comprehend the semantic relationships from different modality  tokens and  generate responses for given instructions.   
    This process can be expressed as:  
    \begin{align}
        \hat{T} = f_{llm}(T),
    \end{align}
    where $\hat{T}=\left(\hat{t}_1, \hat{t}_2, \ldots, \hat{t}_j\right) \in \mathbb{R}^{j \times c}$,  and $\hat{t}_i$ denotes  the predicted $i$-th token, based on the semantics of all previous tokens  $\{t_{<i}\}$. 
    Subsequently,  $\hat{t}_i$ is passed through a linear layer $f_{llm \rightarrow vocab}$  to be mapped into the vocabulary space.  
    A softmax operation is then applied to compute a probability distribution across the vocabulary, with the word of highest probability designated as the prediction $z_i$ for $\hat{t}_i$.
    The process  can be formulated as: 
    \begin{align}
       \tilde{t}_i &=  f_{llm \rightarrow vocab}(\hat{t}_i), \\
       z_i &= arg \max_{w\in vocab}  \text{Softmax}(\tilde{t}_i)[w].
    \end{align}

     As LLMs are primarily trained on text, a perception gap arises when processing non-textual information. 
    Therefore, we adapt PEFT technology LoRA~\cite{hu2021lora} to the  LLM backbone, and also further fine-tune the normalization layers,  preserving learned knowledge and reducing computational overhead.

\begin{table*}[t]
    \centering
    \setlength\tabcolsep{6.4pt}
    \setlength{\abovecaptionskip}{0.1cm}  
    \renewcommand{\arraystretch}{0.95}
        \caption{Generative 3D object classification results on the ModelNet40 test split and Objaverse.
    The accuracy (\%)  under the \textbf{I}nstruction-typed (I) prompt ``What is this?'' and the \textbf{C}ompletion-type (C) prompt ``This is an object of'' are reported. 
    The  \textbf{bold} and \uline{underline} indicate the  best and second best results, respectively.
   }

    \resizebox*{\linewidth}{!}{

\begin{tabular}{lcrrcccccccc}
\toprule
                        &                             &                                                                       &                                                                                 &                         & \multicolumn{3}{c}{ModelNet40}                   & \multicolumn{3}{c}{Objaverse}                    &                           \\ \cline{6-11}
\multirow{-2}{*}{Model} & \multirow{-2}{*}{Reference} & \multirow{-2}{*}{\begin{tabular}[r]{@{}r@{}}LLM \\ Size\end{tabular}} & \multirow{-2}{*}{\begin{tabular}[r]{@{}r@{}}Trainable \\ Params\end{tabular}} & \multirow{-2}{*}{Input} & (I)            & (C)            & Average        & (I)            & (C)            & Average        & \multirow{-2}{*}{Average} \\ \midrule
InstructBLIP-7B~\cite{dai2024instructblip}          & NeurIPS,23                  & 7B                                                                    & 0.20B                                                                            & Single-V. Img.          & 19.53          & 31.48          & 25.51          & 45.00          & 42.00          & 43.50          & 34.50                     \\
InstructBLIP-13B~\cite{dai2024instructblip}         & NeurIPS,23                  & 13B                                                                   & 0.20B                                                                           & Single-V. Img.          & 25.97          & 31.40           & 28.69          & 37.00          & 31.50          & 34.25          & 31.47                     \\
LLaVA-7B~\cite{liu2024visual}                 & NeurIPS,23                  & 7B                                                                    &  7.03B                                                                               & Single-V. Img.          & 39.75          & 39.67          & 39.71          & 49.50          & 50.50          & 50.00          & 44.86                     \\
LLaVA-13B~\cite{liu2024visual}                & NeurIPS,23                  & 13B                                                                   &   13.03B                                                                              & Single-V. Img.          & 37.12          & 36.06          & 36.59          & 53.00          & 50.50          & 51.75          & 44.17                     \\ \midrule
3D-LLM~\cite{hong20243d}                  & NeurIPS,23                  &   13B                                                                    &      -                                                                           & 3D Obj. + Mul.-V. Img.  & -              & -              & -              & 49.00          & 41.50          & 45.25          & 45.25                     \\
Point-Bind LLM~\cite{guo2023point}          & arXiv,23.9                  & 7B                                                                    &     -                                                                            & 3D Point Cloud          & 51.90          & 39.71          & 45.81          & 6.00           & 4.50           & 5.25           & 25.53                     \\
PointLLM-7B~\cite{xu2023pointllm}              & arXiv,23.8                  & 7B                                                                    & 7.01B                                                                           & 3D Point Cloud          & {\ul 53.44}          & 51.82          & 52.63          & 55.00          & 51.00          & 53.00          & 52.82                     \\
PointLLM-13B~\cite{xu2023pointllm}             & arXiv,23.8                  & 13B                                                                   & 13.01B                                                                          & 3D Point Cloud          & 53.00          & {\ul 52.55}          & 52.78          & {\ul 56.50}          & {\ul 51.50}          & 54.00          & 53.39                     \\
ShapeLLM-7B~\cite{qi2024shapellm}              & arXiv,24.2                  & 7B                                                                    & 7.04B                                                                           & 3D Point Cloud          & -              & -              & {\ul 53.08}          & -              & -              & {\ul 54.50}          & {\ul 53.79}                     \\
ShapeLLM-13B~\cite{qi2024shapellm}             & arXiv,24.2                  & 13B                                                                   & 13.04B                                                                          & 3D Point Cloud          & -              & -              & 52.96          & -              & -              & 54.00          & 53.48                     \\
\rowcolor[HTML]{EFEFEF} 
\cellcolor[HTML]{EFEFEF}     & \cellcolor[HTML]{EFEFEF}                 & \cellcolor[HTML]{EFEFEF}                                                      & \textbf{0.05B}                                                              & \cellcolor[HTML]{EFEFEF} & \textbf{61.75} & \textbf{59.97} & \textbf{60.86} & \textbf{60.00} & \textbf{60.50} & \textbf{60.25} & \textbf{60.56}    
\\
\rowcolor[HTML]{EFEFEF} 
\multirow{-2}{*}{\cellcolor[HTML]{EFEFEF}\textbf{MiniGPT-3D}}     & \multirow{-2}{*}{\cellcolor[HTML]{EFEFEF}\textbf{-}}                  & \multirow{-2}{*}{\cellcolor[HTML]{EFEFEF}\textbf{2.7B}}                                                         & { \textbf{(47.8M)}}                                                                & \multirow{-2}{*}{\cellcolor[HTML]{EFEFEF}\textbf{3D Point Cloud}}    & {\color[HTML]{009901} \textbf{(+8.31)}} & {\color[HTML]{009901} \textbf{(+7.42)}}  & {\color[HTML]{009901} \textbf{(+7.78)}} & {\color[HTML]{009901} \textbf{(+3.5)}} & {\color[HTML]{009901} \textbf{(+9.00)}} & {\color[HTML]{009901} \textbf{(+5.75)}}  & {\color[HTML]{009901} \textbf{(+6.77)}}          \\ \bottomrule
\end{tabular}
}
    \label{tab:class}

\end{table*}

\begin{table*}[t]
  \centering
  \setlength\tabcolsep{7.9pt}
  \setlength{\abovecaptionskip}{0.1cm}  
  \renewcommand{\arraystretch}{0.95}
  
  \caption{3D object captioning results on Objaverse. The results are from  human evaluation, GPT-4 evaluation, and traditional metrics. 
  The  \textbf{bold} and \uline{underline} indicate the  best and second best results, respectively.
   }

  \resizebox*{\linewidth}{!}{
\begin{tabular}{lcrrcccccc}
\toprule
\multirow{2}{*}{Model} & \multirow{2}{*}{Reference} & \multirow{2}{*}{\begin{tabular}[c]{@{}r@{}}LLM\\ Size\end{tabular}} & \multirow{2}{*}{\begin{tabular}[c]{@{}r@{}}Trainable\\ Params\end{tabular}} & \multirow{2}{*}{GPT-4} & \multirow{2}{*}{Sentence-BERT} & \multirow{2}{*}{SimCSE} & \multicolumn{3}{c}{Human Evaluation}  \\ \cline{8-10} 
   &  &   &   &  &  &   & Correctness & Hallucination \textbf{↓} & Precision \\ \midrule
InstructBLIP-7B~\cite{dai2024instructblip}   & NeurIPS,23   & 7B  & 0.20B  & 45.34 & 47.41 & 48.48 & 2.56  & 0.77  & 76.99 \\
InstructBLIP-13B~\cite{dai2024instructblip}   & NeurIPS,23   & 13B   & 0.20B   & 44.97 & 45.90 & 48.86 & 2.58  & 1.13  & 69.56\\
LLaVA-7B~\cite{liu2024visual}  & NeurIPS,23   & 7B  &  7.03B  & 46.71 & 45.61 & 47.10 & 2.76  & 0.86  & 76.30  \\
LLaVA-13B~\cite{liu2024visual}  & NeurIPS,23   & 13B   &   13.03B   & 38.28 & 46.37 & 45.90 & 2.43  & 0.86  & 73.97 \\ \midrule
3D-LLM~\cite{hong20243d}  & NeurIPS,23   & 13B  &  -   & 33.42 & 44.48 & 43.68  & 1.77  & 1.16  & 60.39\\
PointLLM-7B~\cite{xu2023pointllm}   & arXiv,23.8   & 7B  & 7.01B  & 44.85 & 47.47 & 48.55 & 3.04  & \textbf{0.66}  & {\ul 82.14} \\
PointLLM-13B~\cite{xu2023pointllm}   & arXiv,23.8   & 13B   & 13.01B  & 48.15 & 47.91 & 49.12  & {\ul 3.10}  & 0.84  & 78.75 \\
ShapeLLM-7B~\cite{qi2024shapellm}   & arXiv,24.2   & 7B  & 7.04B  & 46.92 & 48.20  & 49.23  & -   & -   & -  \\
ShapeLLM-13B~\cite{qi2024shapellm}   & arXiv,24.2   & 13B   & 13.04B & {\ul 48.94}   & {\ul 48.52}   & {\ul 49.98}  & -   & -   & -  \\
\rowcolor[HTML]{EFEFEF} 
\cellcolor[HTML]{EFEFEF}  & \cellcolor[HTML]{EFEFEF}   & \cellcolor[HTML]{EFEFEF}  & \textbf{0.05B}  & \textbf{57.06}  & \textbf{49.54}  & \textbf{51.39}  & \textbf{3.50}   & {\ul \textbf{0.71}}   & \textbf{83.14} \\
\rowcolor[HTML]{EFEFEF} 
\multirow{-2}{*}{\cellcolor[HTML]{EFEFEF}\textbf{MiniGPT-3D}} & \multirow{-2}{*}{\cellcolor[HTML]{EFEFEF}\textbf{-}} & \multirow{-2}{*}{\cellcolor[HTML]{EFEFEF}\textbf{2.7B}} &  \textbf{(47.8M)}   & {\color[HTML]{009901} \textbf{(+8.12)}} & {\color[HTML]{009901} \textbf{(+1.02)}} & {\color[HTML]{009901} \textbf{(+1.41)}}  & {\color[HTML]{009901} \textbf{(+0.40)}} & {\color[HTML]{996671} \textbf{(+0.05)}} & {\color[HTML]{009901} \textbf{(+1.00)}} \\ \bottomrule
\end{tabular}
}
   \label{tab:caption}

\end{table*}

\subsection{Training Stages}
    \label{Training_Stages}

    To gradually transfer the priors of 2D-LLM to point cloud modality and enhance the nascent 3D-LLM's comprehension, our training process includes four  stages, each focusing on a distinct task, as shown in Figure~\ref{arch}.
    The following subsections will describe them.

\subsubsection{\textbf{Stage I}}
    
    As shown in Figure~\ref{arch}(a), the first stage aims to bridge the knowledge gap between the 3D point cloud encoder and 2D-LLM modules, facilitating a seamless transition from 3D to 2D.
    We solely train the point cloud projection layer (MLP), with   other modules   frozen.
    Initialization is sourced from ULIP-2~\cite{xue2023ulip} for the encoder, BLIP-2~\cite{li2023blip} for Q-Former, and TinyGPT-V~\cite{yuan2023tinygpt} for normalization layers of LLM, LoRA of LLM, and the modality projector. 
    Since the frozen Q-Former from BLIP-2 is also used in TinyGPT-V, MiniGPT-3D  only owns two knowledge domains  from    3D  of  ULIP-2 and  2D-LLM of TinyGPT-V    before training.
    To build a robust bridge between domains, we train the projection layer using 660k caption-point cloud pairs, involving 1.4M parameters, as detailed in Table~\ref{all_stages}.

\subsubsection{\textbf{Stage II}}
    In the second stage, our objective is to transfer the vision-language knowledge domain to 3D space, establishing the 3D-language knowledge domain.
    As shown in Figure~\ref{arch}(b), we fine-tune four parts: the point cloud projection layer (MLP), the Q-Former, the modality projector, and the LLM.
    Utilizing the  3D-2D bridge of the first stage, 2D-LLM modules, via fine-tuning, gain comprehension of 3D point clouds and gradually   transfer  the powerful priors to be the 3D-language knowledge.
    During this process, to minimize the impact of the 3D-2D bridge, we employ the identical dataset from the first stage to train 47.4M parameters, as outlined in Table~\ref{all_stages}.
    
\subsubsection{\textbf{Stage  III}}
 
    To gain better 3D-language knowledge, we further fine-tune  the modules trained in the second stage and utilize a more challenging dataset, including detailed caption-point cloud pairs and conversations, to empower MiniGPT-3D with the capabilities to comprehend and respond to complex instructions.

\subsubsection{\textbf{Stage   IV}}

    During the prior   stages,    using a single set of   queries  restricts 3D perception perspective, leading to incomplete  cognition.
    To  refine MiniGPT-3D's perception, we introduce MQE to adaptively activate suitable multiple query experts for Q-Former, as shown in Figure~\ref{arch}(d).
    Distinct from the preceding three stages focusing on rapidly establishing 3D-language knowledge,  this stage presents a stable semantic context  for query experts to learn knowledge efficiently.
    Specifically, we only fine-tune 0.4M MQE-related parameters, reusing the dataset from the third stage to minimize the impact of data distribution changes, as outlined in Table~\ref{all_stages}.
    
\subsection{Training Objective}
    \label{Training_Loss}

    The training objective of MiniGPT-3D aims to minimize the discrepancy between predicted and true probability distributions at each token position.
    Given a point cloud and corresponding text instruction, MiniGPT-3D outputs a sequence $\hat{T}$.
    Next, $\hat{T}$ is processed by $f_{llm \rightarrow vocab}$ and then a softmax operation is applied to obtain the probability distribution over the vocabulary  for each output token, denoted as $\overline{T}$.
    The training loss is formulated as follows:
    \begin{align}
       \mathcal{L} &= \text{CrossEntropy}\left( h(G), ~\overline{T} \right),
    \end{align}
    where the $h(\cdot )$ represents   the LLM's tokenizer. 
    $G$ is the ground truth text.
    The $CrossEntropy(\cdot )$ refers to the cross-entropy loss function. 
    Notably, we only compute the loss for the generated text.

\section{Experiments}

\subsection{Experimental Settings}
    Utilizing  one RTX 3090 GPU with  24GB of RAM, we train MiniGPT-3D with only 47.8M trainable parameters in  26.8 hours. 
    We adopt the AdamW optimizer with a weight decay of 0.05 and a cosine decay with linear warm up learning rate schedule.
    The initial learning rate decreases gradually as the training stage advances, as shown in  Table~\ref{all_stages}.
    We use the point-text instruction dataset~\cite{xu2023pointllm}, including 660K brief-description instructions and 70K complex instructions.
    200 objects  are splited as test data, following PointLLM~\cite{xu2023pointllm} and ShapeLLM~\cite{qi2024shapellm}.
    For each input point cloud $P\in\mathbb{R}^{n\times d}$, the  number of point $n$ is 8192, and the dimension $d$ is 6.
    We default point clouds without color to black.
    For a fair comparison, we adopt the identical versions models of GPT-4~\cite{openai2023gpt} (``gpt-4-0613'') and ChatGPT~\cite{ChatGPT} (``gpt-3.5-turbo-0613'') as our evaluation tools, like prior works~\cite{xu2023pointllm, qi2024shapellm}.
    We choose multiple SOTA 3D-LLMs~\cite{ hong20243d,guo2023point,xu2023pointllm,qi2024shapellm} and two popular open-source 2D-LLMs~\cite{dai2024instructblip, liu2024visual} as our baselines.

\begin{table*}[t]
\small
\centering
\setlength{\abovecaptionskip}{0.1cm}  
\caption{\textbf{Qualitative comparisons.} The classification and caption results of models on ModelNet40 and Objaverse are presented. Our MiniGPT-3D generates more detailed and insightful responses compared to other baselines.}

\label{tab:demo}

    \centering
    \renewcommand{\arraystretch}{1}

    \resizebox*{\linewidth}{!}{
\begin{tabular}{@{}l p{0.48\linewidth} p{0.48\linewidth}}
\toprule

Samples 1, 2 & 
  \begin{minipage}{\linewidth}
    \hspace{1.6em}
    \includegraphics[width=0.16\linewidth,height=0.12\linewidth]{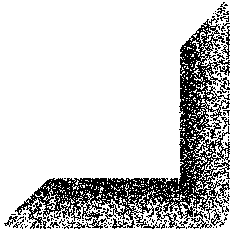}
    \hspace{1.6em}
    \includegraphics[width=0.16\linewidth,height=0.12\linewidth]{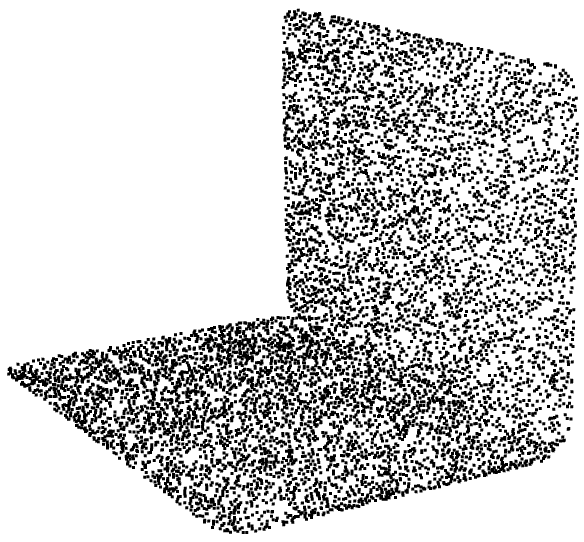}
  \end{minipage}
& 
  \begin{minipage}{\linewidth}
    \hspace{1.6em}
    \includegraphics[width=0.20\linewidth,height=0.12\linewidth]{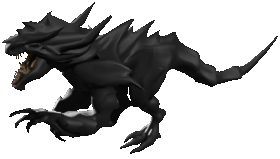}
    \hspace{1.6em}
    \includegraphics[width=0.20\linewidth,height=0.12\linewidth]{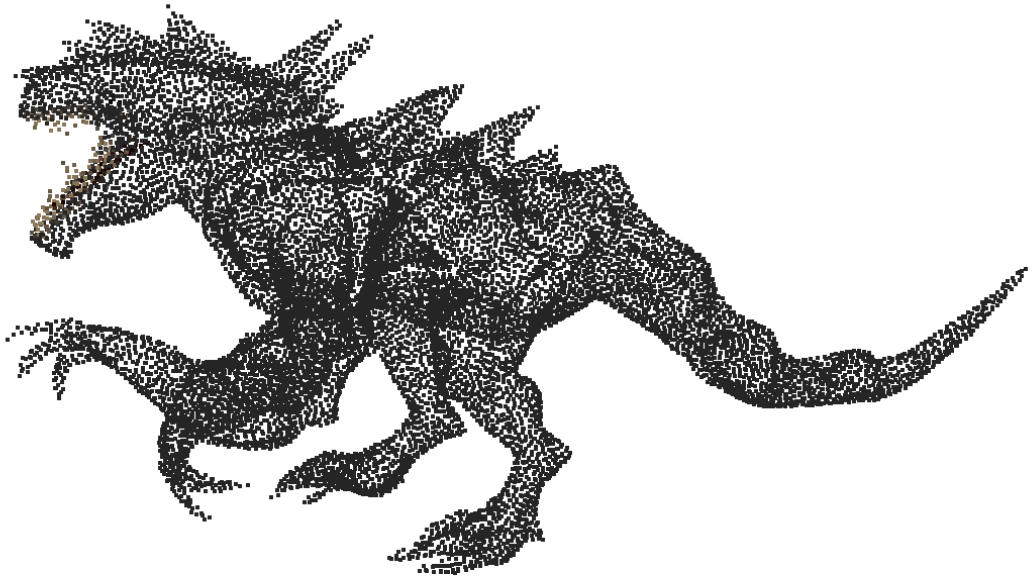}
  \end{minipage}
\\ \midrule
Prompt & What is this? & Briefly caption this 3D model. \\  
Ground Truth & Laptop & A cartoon black monster like a dragon \\  
\midrule

InstructBLIP-13B~\cite{dai2024instructblip} & symbol letter l & a black lizard with a sharp tooth in a dark room \\ 
LLaVA-13B\cite{liu2024visual}  & A small, grainy, black and white letter j. & A 3D model of a dark, menacing dragon. \\ 
3D-LLM~\cite{hong20243d} & - & A black and white tiger with long legs, standing on its hind leg. \\
Point-Bind LLM~\cite{guo2023point}  & This is a laptop computer. & The 3D model features a large, ornate gargoyle with a horned helmet, sitting on top of a building. \\
PointLLM-13B~\cite{xu2023pointllm} & The 3D model represents a notebook computer, typically a laptop. & The 3D model depicts a menacing black dragon, with its mouth opened wide revealing a row of sharp teeth. \\

\textbf{MiniGPT-3D} &  \textbf{This is a 3D model of a sleek, black {\color{LightGreen}laptop}. The laptop is {\color{LightGreen}rectangular} in shape and has a glossy finish, suggesting a high-quality build. It has {\color{LightGreen}a flat screen}, which is a common feature in modern laptops, and it's equipped with {\color{LightGreen}a keyboard and a touchpad} for user interaction. The laptop appears to be in good condition, suggesting it could be used for various tasks such as {\color{LightGreen}work, entertainment, or education.}} & \textbf{This is a 3D model of a fearsome, {\color{LightGreen}black-colored dinosaur}. The dinosaur possesses an {\color{LightGreen}aggressive stance}, with {\color{LightGreen}sharp claws} extending from its arms and legs. It has long black fangs that {\color{LightGreen}seem ready to attack.}} \\
\bottomrule

\end{tabular}
}
\end{table*}
\begin{table*}[t] 
  \begin{minipage}[t]{0.33\linewidth}  
    \centering
    \setlength\tabcolsep{3pt}
    \setlength{\abovecaptionskip}{0.1cm}
    \renewcommand{\arraystretch}{0.95}
            \caption{Ablation on  training process.
           }

            \resizebox*{\linewidth}{!}{
        \begin{tabular}{c|ccccc}
        \toprule
        Row No. & Stage  I & Stage II & Stage III & Stage IV & Acc. \\ \midrule
         1  & $\checkmark$ &                      &                      &                      & 39.10                     \\
         2  & $\checkmark$ & $\checkmark$ &                      &                      & 55.92                     \\
         3  & $\checkmark$ & $\checkmark$ &   $\checkmark$                   &                      & 59.10                     \\
        \rowcolor[HTML]{EFEFEF} 
         4  & $\checkmark$ & $\checkmark$ & $\checkmark$ & $\checkmark$ & \textbf{60.56}                     \\
         5  & $\checkmark$ &          & $\checkmark$ & $\checkmark$ & 52.81                     \\
        6  &          & $\checkmark$ & $\checkmark$ & $\checkmark$ & 58.46                     \\
        7   & $\checkmark$ & $\checkmark$ &         & $\checkmark$ &    47.93                   \\
        \bottomrule
        \end{tabular}
        }
           \label{tb:ablation_train_progress}
        
    \vfill
  \end{minipage}%
  \hfill 
  \begin{minipage}[t]{0.35\linewidth}  
  \setlength{\abovecaptionskip}{0.1cm}
    \centering
    \renewcommand{\arraystretch}{0.95}
    \caption{Ablation on 2D priors  from 2D-LLM.}
          \resizebox*{0.66\linewidth}{!}{
    \begin{tabular}{ccc}
    \toprule
    \begin{tabular}[c]{@{}c@{}}Modality\\ Projector\end{tabular} & \begin{tabular}[c]{@{}c@{}}Norm and\\LoRA  for LLM\end{tabular}   & Acc.   \\ \midrule
      &     & 49.04 \\ 
        $\checkmark$      &              & 57.44 \\
        &   $\checkmark$        & 57.86 \\
          \rowcolor[HTML]{EFEFEF} 
        $\checkmark$      &   $\checkmark$        & \textbf{58.46} \\
       \bottomrule
    \end{tabular}
        }  
        \label{tab: 2D_prior}

    \vfill
  \end{minipage}
     \hfill 
  \begin{minipage}[t]{0.3\linewidth}  
    \setlength{\abovecaptionskip}{0.1cm}
     \centering
     \renewcommand{\arraystretch}{0.95}
    \caption{Ablation on  stages using MQE.}
     \setlength\tabcolsep{3pt}
     \resizebox*{0.9\linewidth}{!}{
        \begin{tabular}{ccccc}
        \toprule
        Stage I & Stage II & Stage III & Stage IV & Acc. \\ \midrule
         $\checkmark$             & $\checkmark$             & $\checkmark$            & $\checkmark$            & 58.83                      \\
                      & $\checkmark$             & $\checkmark$            & $\checkmark$            & 60.25                      \\
                       &               & $\checkmark$            & $\checkmark$            & 59.50                      \\
        \rowcolor[HTML]{EFEFEF} 
                       &               &              & $\checkmark$            & \textbf{60.56}             \\ \bottomrule
        \end{tabular}
        }
        \label{tab: ablation_stage_MQE}
\vfill
  \end{minipage}
  
\end{table*}

\subsection{Generative 3D Object Classification}
 We conduct the generative 3D object classification tasks~\cite{xu2023pointllm} on ModelNet40~\cite{wu20153d}  and Objaverse~\cite{deitke2023objaverse} datasets to assess MiniGPT-3D's categorical cognitive ability.
 
    \paragraph{\textbf{Settings}}
        For a fair comparison, we utilize the classification evaluation settings similar to prior works~\cite{xu2023pointllm,qi2024shapellm}.
        We employ  identical prompts: the \textbf{I}nstruction-typed (I) prompt ``What is this?'' and the \textbf{C}ompletion-type (C) prompt ``This is an object of''.
        Point clouds and these prompts are fed into our MiniGPT-3D, outputting  textual responses.
        For close-set zero-shot classification on ModelNet40, ChatGPT processes  the text responses of MiniGPT-3D   to select predicted categories from 40  ModelNet40 classes.
        For open-vocabulary classification on Objaverse, GPT-4 is employed as an evaluator to determine whether  MiniGPT-3D's text response refers to the same category as the ground-truth caption. 

    \paragraph{\textbf{Results}}
        Experimental results are   shown in Table~\ref{tab:class}.
        We achieve  SOTA performance on all classification benchmarks using  only one RTX 3090.
        Specifically, compared to the best baseline, ShapeLLM~\cite{qi2024shapellm}, we achieve significant improvements of 7.78\% and 5.75\% in average accuracy on ModelNet40 and Objaverse datasets, respectively.
        Unlike other methods using LLM (7B or 13B) that require  fine-tuning on  8 A100 or 8 A800  for hundreds of total GPU-hours,  our  MiniGPT-3D  only utilizes a 2.7B LLM and trains 47.8M parameters on a single RTX 3090 GPU in 27 hours.
        These   demonstrate the superiority and efficiency of our MiniGPT-3D, which leverages the powerful priors from 2D-LLMs to build 3D-LLM.
        Additionally, we observe that MiniGPT-3D exhibits the best performance in recognizing   3D objects of the ModelNet40 dataset unused during training, indicating its stronger generalization ability compared to other methods. 
        Furthermore, even if using different prompts ( \textbf{I} and \textbf{C}) on the Objaverse dataset, MiniGPT-3D demonstrates highly consistent classification performance compared to other 3D-LLMs that have  a 4\% accuracy gap, showcasing its robustness for diverse prompts.

\subsection{3D Object Captioning}
To assess the model's understanding of 3D object details, we perform the 3D object captioning task.

    \paragraph{\textbf{Settings}}
        For a fair comparison, we follow the evaluation settings of  prior works~\cite{xu2023pointllm,qi2024shapellm}.
        We use the  prompt ``Caption this 3D model in detail''.
        We adopt three distinct evaluation methods: human evaluation, GPT-4~\cite{openai2023gpt} evaluation, and traditional metric evaluation. 
        In human evaluation, volunteers  evaluate the model using standardized processes from PointLLM~\cite{xu2023pointllm}.
        Specifically, focusing on object attributes (such as type, color, material, etc.), volunteers visually assess objects and assign correctness scores and hallucination scores to captions.
        Correctness measures model accuracy in describing attributes, while hallucination evaluates fabricated details' severity.
        Each attribute, correct or hallucinated, receives a point. Precision is calculated as the ratio of correct information in model-generated content.
        The Inter-Annotator Agreement score is 0.89 on ICC1k, indicating volunteers' high consistency in cognitive understanding and scoring criteria.
         GPT-4 evaluates  semantic similarity between our model's output and manually annotated captions.
        In traditional metric evaluation,  like prior works~\cite{xu2023pointllm,qi2024shapellm}, we use data-driven metrics like Sentence-BERT~\cite{reimers2019sentence} and SimCSE~\cite{gao2021simcse}, instead of BLEU-1~\cite{papineni2002bleu}, ROUGEL~\cite{lin2004rouge}, and METEOR~\cite{banerjee2005meteor},  because   the latter lack sufficient capabilities in semantic evaluation.

    \paragraph{\textbf{Results}}
        As shown in Table~\ref{tab:caption}, our MiniGPT-3D achieves SOTA performance on  multiple  metrics.
        Specifically, MiniGPT-3D outperforms ShapeLLM-13B~\cite{qi2024shapellm}, by a large margin of 8.12  on the GPT-4 evaluation score, setting new SOTA with only 2.7B LLM,  indicating robust 3D detail comprehension.
        Also, compared to ShapeLLM-13B, MiniGPT-3D surpasses 1.02 and 1.41 on Sentence-BERT and SimCSE metrics, respectively, achieving new SOTA with its remarkable ability to generate  accurate captions matching ground truth.
        Human evaluation further reveals  MiniGPT-3D's superior correctness and precision scores compared to baselines.
        Notably, even with a 2.7B LLM, MiniGPT-3D exhibits a hallucination score comparable to SOTA, surpassing larger 13B LLM-based methods.
        These  outstanding results showcase MiniGPT-3D's fine-grained understanding of 3D objects, inheriting the cognitive capabilities of 2D-LLM.

\vspace*{-0.8mm}

\subsection{Qualitative Results}
    Figure~\ref{fig:teaser}(e) qualitatively shows the MiniGPT-3D's powerful ability to perceive 3D object details.
    Our MiniGPT-3D precisely extracts information from 3D objects, encompassing categories, colors, shapes, materials, and internal component relationships.
    Additionally, MiniGPT-3D  can perform reasonable reasoning  based on object cues, such as potential occurrence periods and locations.
    Figure~\ref{fig:teaser}(f) further demonstrates MiniGPT-3D's comprehension of 3D object information in open-ended dialogues.
    MiniGPT-3D accurately outputs 3D object-related world knowledge, showcasing its  extensive textual knowledge inherited from LLMs.    
    
    In sample 1 of Figure~\ref{tab:demo},  our MiniGPT-3D successfully recognizes the shape, screen, and keyboard of a laptop, compared to other methods.
    Furthermore, it can deduce the potential usage of this 3D object.  
    In the more complex sample 2 of Figure~\ref{tab:demo}, our MiniGPT-3D demonstrates superior understanding capabilities of 3D objects by recognizing additional features like the dinosaur's sharp claws and inferring its potential action intentions, compared to other methods.

\vspace*{-0.8mm}

\subsection{Ablation Studies}
    In this section, we conduct ablation studies to investigate various model design options. Herein, we report the total average accuracy of MiniGPT-3D on the generative classification benchmark.

    \subsubsection{ \textbf{Training process} } We conduct ablation study to validate the efficacy of our four-stage training strategy.
    The results in  Table~\ref{tb:ablation_train_progress} highlight the optimal performance achieved by our approach.
    Specifically, comparing Row \#4 vs.  \#6, we observe that the first stage bridges knowledge between 2D-LLM and 3D encoder, enabling smoother semantic transitions across different dimensional spaces.
    Comparing Row \#4  vs.  \#5, we note that  the second training stage  which involves using easy tasks to adapt the knowledge of the 2D-LLM to the 3D space, allows the model to focus on enhancing cognitive capabilities in subsequent stages.
    Comparing Row \#4  vs.  \#7,  the third training stage  utilizes more challenging tasks to reinforce the newborn 3D cognitive abilities, providing a reliable semantic context for the final stage  to train MQE.
    Comparing Row \#4 vs.  \#3, the inclusion of the fourth  stage, dedicated to training the MQE, enables each query expert to acquire unique knowledge, further enhancing MiniGPT-3D's understanding of 3D objects.

\begin{table}[t] 
    \begin{minipage}[t]{0.5\linewidth}  
     \centering
     \setlength{\abovecaptionskip}{0.1cm}
     \renewcommand{\arraystretch}{0.86}
    \caption{Ablation on fine-tuned modules in Q-Former.}
        \resizebox*{\linewidth}{!}{
            \begin{tabular}{cccc}
            \toprule
            LoRA & LoRA & \multirow{2}{*}{Norm} & \multirow{2}{*}{Acc.} \\ 
             
             Q, K, V & Dense     &                       &                           \\ \midrule
                            &         &                          & 58.18                      \\
             $\checkmark$              &         &                          & 59.85                      \\
             $\checkmark$              & $\checkmark$       &                          & 59.97                      \\
              $\checkmark$              & $\checkmark$       & $\checkmark$                        & 60.14                      \\ 
            \rowcolor[HTML]{EFEFEF} 
             $\checkmark$              &         & $\checkmark$                        & \textbf{60.56}                     \\
            \bottomrule
            \end{tabular}
            }

            \label{tab:abl_fine-tuned_modules_of_Q-Former}
        
  \end{minipage}%
     \hfill 
  \begin{minipage}[t]{0.45\linewidth}  
    \centering
    \setlength\tabcolsep{18pt}
    \setlength{\abovecaptionskip}{0.1cm}
    \renewcommand{\arraystretch}{0.905}
    \caption{Ablation on the number of query experts. }
    \resizebox*{\linewidth}{!}{
        \begin{tabular}{cc}
        \toprule
                               &                        \\
        \multirow{-2}{*}{Number} & \multirow{-2}{*}{Acc.} \\ \midrule
        1    & 59.19          \\
        3    & 59.66          \\
        6    & 59.14          \\
        \rowcolor[HTML]{EFEFEF} 
        8    & \textbf{60.56} \\
        10   & 59.85          \\
        \bottomrule
        \end{tabular}
        }
           \label{tb:ablation_num_query_expert}
  \end{minipage}  
  
\vspace*{-2mm}
\end{table}
\begin{table}[t] 
 
  \begin{minipage}[t]{0.5\linewidth}  
    \centering
    \setlength{\abovecaptionskip}{0.1cm}
    \setlength\tabcolsep{12pt}
    \renewcommand{\arraystretch}{0.9}
    \caption{Ablation on the  point  \\ cloud projection layers.}
    \resizebox*{\linewidth}{!}{
        \begin{tabular}{cc}
        \toprule
        Number  of layers  & Acc. \\\midrule
        1                & 57.02          \\
        \rowcolor[HTML]{EFEFEF} 
        2                & \textbf{60.56} \\
        3                & 59.20          \\ \bottomrule
        \end{tabular}
        }
           \label{tab: ablation_pc_linear}
  \end{minipage}
    \hfill 
  \begin{minipage}[t]{0.45\linewidth}  
    \centering
    \setlength\tabcolsep{11pt}
    \setlength{\abovecaptionskip}{0.1cm}
    \renewcommand{\arraystretch}{0.955}
    \caption{Ablation   on router type of MQE.}
    \resizebox*{\linewidth}{!}{
    \begin{tabular}{cc}
    \toprule
    Type     & Acc.       \\ \midrule
    Constant Router & 60.10          \\
    \rowcolor[HTML]{EFEFEF} 
    Sparse Router   & \textbf{60.56} \\
    Soft Router     & 60.31          \\ \bottomrule
    \end{tabular}
    }
       \label{tab: ablation_routet_type}
  \end{minipage}  
\vspace*{-2mm}
\end{table}
\begin{table}[t]
    \centering
    \setlength\tabcolsep{5pt}
    \setlength{\abovecaptionskip}{0.1cm}  

    \renewcommand{\arraystretch}{0.9}
    
\caption{Ablation on trained modules in stage IV.
   }
\resizebox*{\linewidth}{!}{
\begin{tabular}{cccccc}
\toprule
MQE & \multicolumn{1}{c}{\begin{tabular}[c]{@{}c@{}}Norm. \& LoRA\\ for Q-Former\end{tabular}} & \begin{tabular}[c]{@{}c@{}}Modality \\ Projector\end{tabular} & \begin{tabular}[c]{@{}c@{}}Norm. \& LoRA\\ for LLM\end{tabular} & MLP & Acc. \\ \midrule

$\checkmark$        & $\checkmark$           & $\checkmark$         & $\checkmark$      & $\checkmark$   & 58.93                      \\ 
$\checkmark$   & $\checkmark$        & $\checkmark$         & $\checkmark$            &     & 59.93                      \\
$\checkmark$      & $\checkmark$     & $\checkmark$         &        &     & 59.02                      \\
$\checkmark$     & $\checkmark$       &           &      &     & 59.64                      \\
\rowcolor[HTML]{EFEFEF} 
$\checkmark$     & &           &    &     & \textbf{60.56}             \\
\bottomrule
\end{tabular}
}
   \label{tab: ablation_traiend_module_stage4}

     \vspace*{-1mm}
\end{table}

    \subsubsection{\textbf{2D priors from 2D-LLM}}   We conduct ablation study to varify the effectiveness of the 2D priors from 2D-LLM, as detailed in Table~\ref{tab: 2D_prior}.
    Since dropping any pre-trained weights of 2D-LLM would make the first training stage infeasible, all cases of this ablation study  are just trained through stages II to IV.
    We find that removing any of 2D-LLM weight degrades performance, and discarding more pre-trained weights of 2D-LLM  causes an up to 9.4\% accuracy drop.
    These results highlight the crucial role of 2D-LLM knowledge in boosting 3D-LLM performance. 
    Using 2D-LLM modules facilitates cost-efficient training of 3D-LLM even on consumer GPUs like RTX 3090 GPU, enhancing accessibility for the community.

    \subsubsection{\textbf{Training stages using MQE}}  We further investigate  the impact of  training MQE in different stages, with detailed results presented in Table~\ref{tab: ablation_stage_MQE}. 
    Our results indicate that introducing MQE in only    stage IV  achieves optimal performance. 
    The I-III stages enable the model to learn enough semantic features, paving the way for MQE to adaptively select useful information in stage IV.

    \subsubsection{\textbf{Fine-tuned modules in Q-Former}}   Employing PEFT  methods to fine-tune Q-Former   can better align  point features with LLM, avoiding expensive computation.
    As outlined in Table~\ref{tab:abl_fine-tuned_modules_of_Q-Former},  fine-tuning the Query, Key, and Value layers  with  LoRA~\cite{hu2021lora}, along with normalization layers, maximizes the potential of Q-Former. 
    Notely, we efficiently fine-tune the 105M-parameter Q-Former using only 0.7M parameters, achieving a 2.38\% accuracy improvement compared to  the frozen Q-Former.

    \subsubsection{\textbf{Number  of query experts}} Within MQE, each query expert holds unique knowledge, facilitating    extraction of point cloud features. 
    Our experiments, in Table~\ref{tb:ablation_num_query_expert},  reveal that 8 query experts yield optimal performance.
    Insufficient experts may compromise information extraction, while excessive ones may affect cooperation among experts.
    Notably, single-expert, i.e. without MQE, results in a 1.37\% accuracy drop, highlighting the superiority of MQE.

    \subsubsection{\textbf{Point cloud projection layer}}  The point cloud projection layer  bridges point cloud features with the  2D semantics of frozen Q-Former, while ensuring dimensional alignment. 
    As shown in Table~\ref{tab: ablation_pc_linear}, our experiments  demonstrate that two MLP layers offer the optimal setup, as excessive or insufficient layers can result in information loss, compromising overall performance.

    \subsubsection{\textbf{Router type of MQE}} The routing mechanism in MQE regulates the cooperation among query experts.
    The constant router~\cite{kudugunta2021beyond} assigns static average weights, while the soft router~\cite{puigcerver2023sparse} dynamically assigns weights during training.
    The sparse router~\cite{shazeer2017outrageously} selects the top two experts based on the dynamic weights provided by the soft router.
    We explore these router types in Table~\ref{tab: ablation_routet_type}, finding that the sparse router, which dynamically  assigns weights and selects the most promising experts, maximizes the capabilities of MQE.

    \subsubsection{\textbf{Trained modules in stage IV}}   In the training stage IV, only MQE is trained to enable each query expert to learn knowledge within a stable semantic context.
    Our experiments in Table~\ref{tab: ablation_traiend_module_stage4} investigate the integration of various training modules. 
    The results indicate that stage IV is to adaptively aggregate features of different experts, with knowledge gained from I-III stages frozen. Losing any frozen knowledge causes information loss, demonstrating the MQE is specifically designed for information aggregation.

\section{Conclusion}
\label{sec:conclusion}
In this paper, we  present MiniGPT-3D, a efficient and powerful 3D-LLM,   requiring the training of only 47.8M  learnable  parameters within 26.8 hours on one single NVIDIA RTX 3090 GPU.
Specifically, we propose a novel four-stage training strategy that gradually  aligns 3D point cloud features with LLM using 2D priors from 2D-LLM.
Additionally, we design the mixture of query experts, introducing MoE to queries, to adaptively aggregate multiple features.
Extensive experiments show the superiority of MiniGPT-3D in 3D point cloud understanding and question answering.

\textbf{Discussion.} MiniGPT-3D's limitations lie in its training on object-level datasets, preventing it from understanding large-scale point clouds. 
Moreover, like existing 3D-LLMs, our MiniGPT-3D solely focuses on comprehending static 3D objects, lacking the capacity to recognize the actions of dynamic objects.
We will extend our 3D-LLM building approach to    autonomous driving  scenarios.

\bibliographystyle{ACM-Reference-Format}
\bibliography{minigpt3d}

\appendix
\clearpage
\section{Appendix}
\subsection{Qualitative Results}
We   present  more  qualitative results of our MiniGPT-3D,  encompassing 3D recognition and captioning, 3D question answering, as well as qualitative comparisons.

\subsubsection{\textbf{3D Recognition and Captioning}}
\begin{figure*}[t]
    \centering
  \includegraphics[]{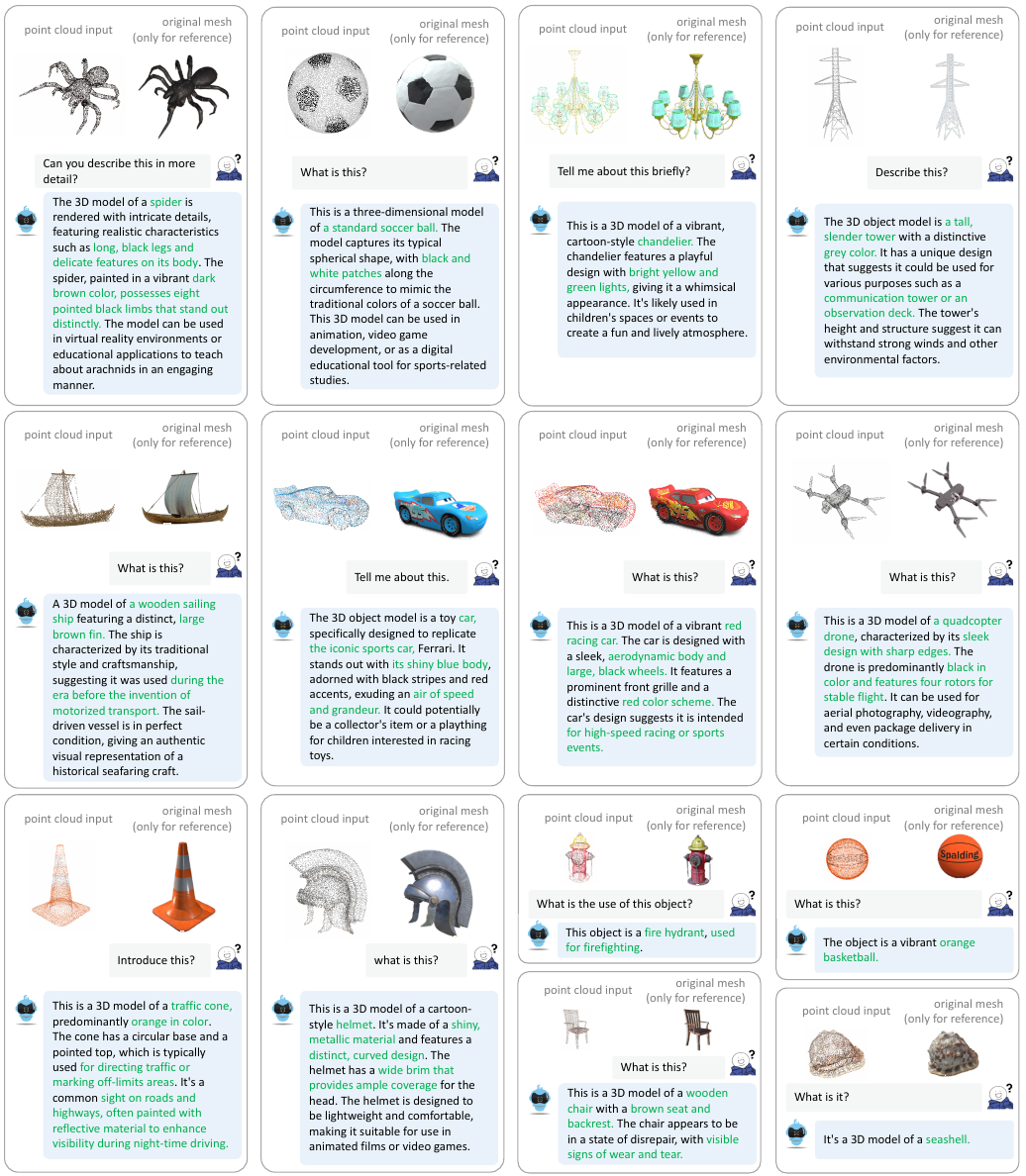}
  \caption{3D recognition and captioning results (\#1) of our MiniGPT-3D. 
  Given a 3D point cloud and a corresponding question, MiniGPT-3D generates a text response.   The correct and fine-grained answers are highlight in  {\color{LightGreen}green}.}
  \label{sup_fig: show_1}
\end{figure*}

\begin{figure*}[t]
    \centering
  \includegraphics[]{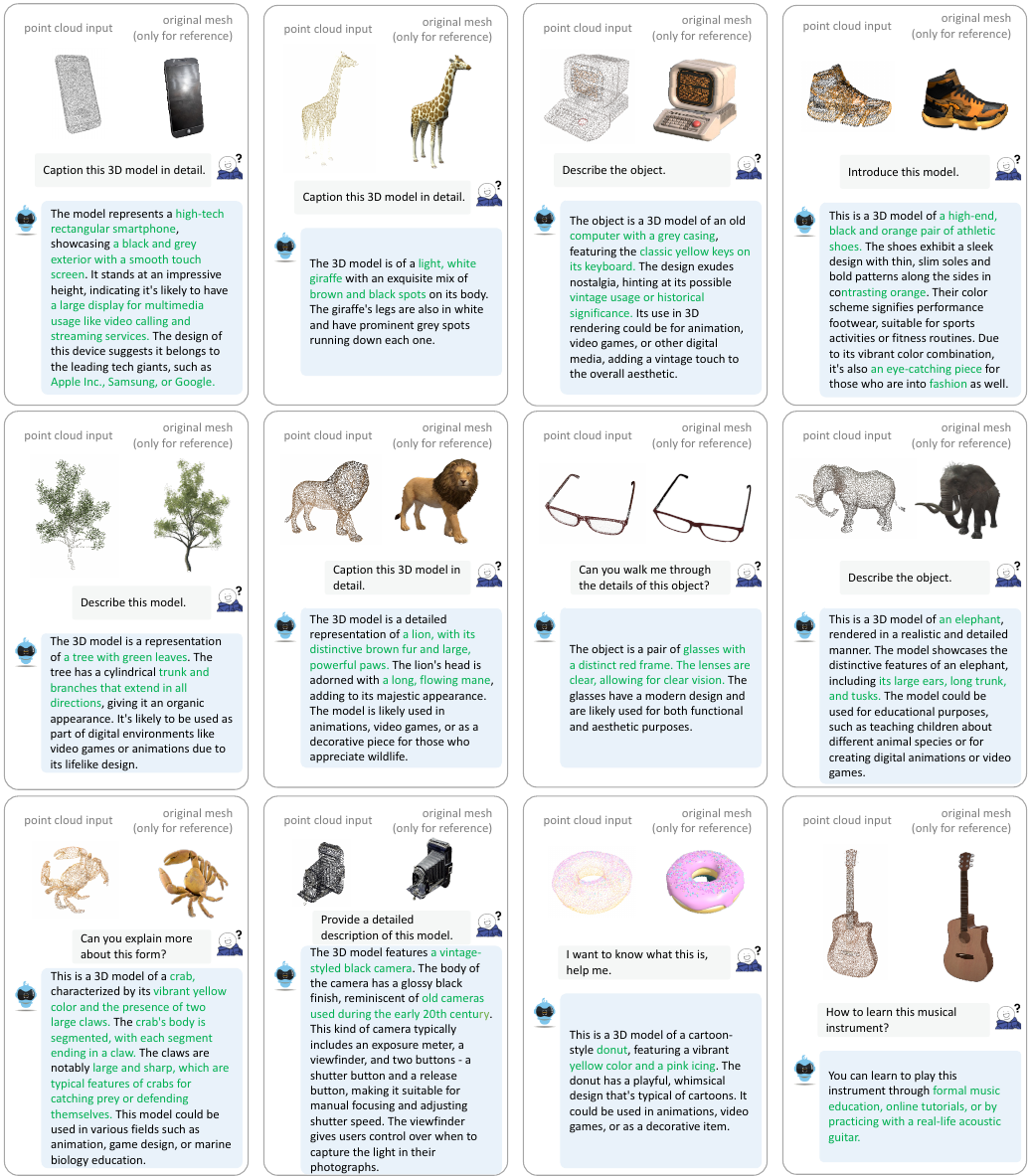}
  \caption{3D recognition and captioning results (\#2) of our MiniGPT-3D. 
  Given a 3D point cloud and a corresponding question, MiniGPT-3D generates a text response.   The correct and fine-grained answers are highlight in  {\color{LightGreen}green}.}
  \label{sup_fig: show_2}
\end{figure*}

Figure~\ref{sup_fig: show_1} and Figure~\ref{sup_fig: show_2} further showcase the qualitative results of our MiniGPT-3D in 3D recognition and captioning.
Given a 3D point cloud and  instruction, MiniGPT-3D is capable of generating text responses that include the object's category, quantity, color, and as well as  unique characteristics.
Furthermore, our MiniGPT-3D also leverages the point cloud information to make   reasonable reasoning, deducing potential uses and emergence timelines. This excellent  comprehension of point clouds underscores the advantage of employing  priors from 2D-LLMs  to build 3D-LLMs.

\subsubsection{\textbf{3D Question Answering}}

\begin{figure*}[t]
    \centering
    \setlength{\abovecaptionskip}{0.15cm} 
  \includegraphics[]{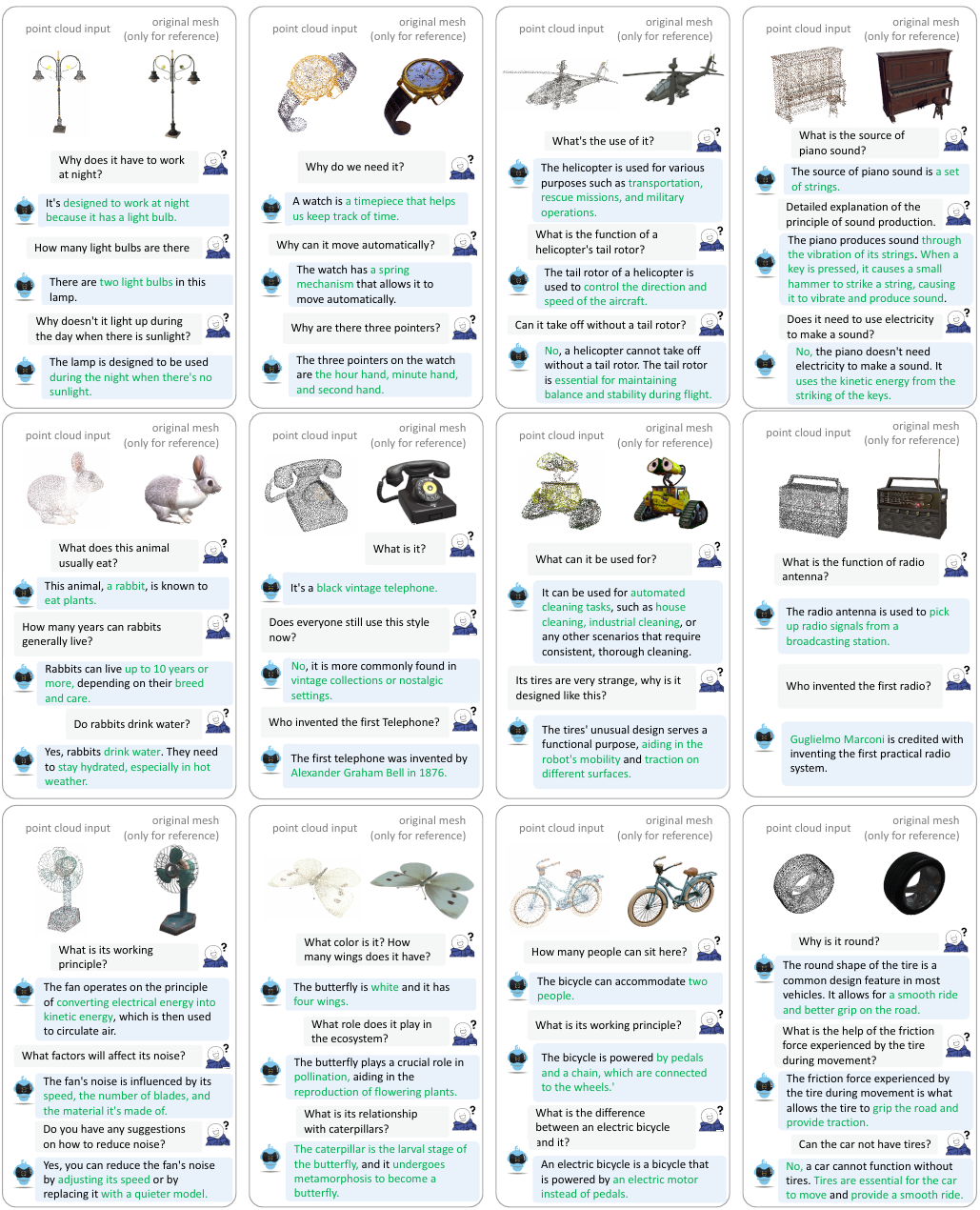}
    \caption{3D question answering results  of our MiniGPT-3D. 
   MiniGPT-3D supports multi-round conversation  regarding the 3D object. 
The correct and fine-grained answers are highlight in  {\color{LightGreen}green}.}
  \label{sup_fig: show_3}
\end{figure*}

Figure~\ref{sup_fig: show_3} further provides the qualitative results of our MiniGPT-3D on 3D question answering.
Our MiniGPT-3D supports multi-turn dialogues with users regarding the input 3D point cloud. Users can continuously pose various open-ended questions to MiniGPT-3D about the 3D object, such as its working principle, the number of objects, specific historical event times, and even logical questions. Despite only training 47.8M trainable parameters on one single NVIDIA  RTX 3090 GPU for 27 hours,  through these examples, we observe that our MiniGPT-3D possesses extensive general knowledge and maintains contextual coherence in multi-turn dialogues, outputting correct text  responses.
These impressive results underscore the superiority of efficiently aligning 3D point clouds with LLMs based on 2D-LLM knowledge.

\subsubsection{\textbf{Qualitative comparisons}}

We present more qualitative comparisons, similar to Table 4 in our
main paper.   The results are shown in Table~\ref{app_tab:demo_modelnet40_1}.
Compared with other methods, our MiniGPT-3D outputs a more detailed text response, while accurately recognizing object categories and capturing more 3D point cloud information, such as usage, shape, internal components, geometric attributes, materials, etc.
The  results show the excellent point cloud understanding capabilities  of   MiniGPT-3D.

\begin{table*}[t]
\small
\centering
\caption{ Qualitative results on ModelNet40. The first image in each sample serves as the input for image-based models, while additional views are provided for reference purposes.
Compared to other methods, MiniGPT-3D demonstrates superior accuracy in identifying object categories, capturing richer information, and generating more comprehensive responses.
}

\label{app_tab:demo_modelnet40_1}
\begin{NiceTabular}{@{}l p{0.38\linewidth} p{0.38\linewidth}}
\toprule

Samples 1, 2 & 
  \begin{minipage}{\linewidth}  
    \includegraphics[width=0.30\linewidth,height=0.15\linewidth]{ 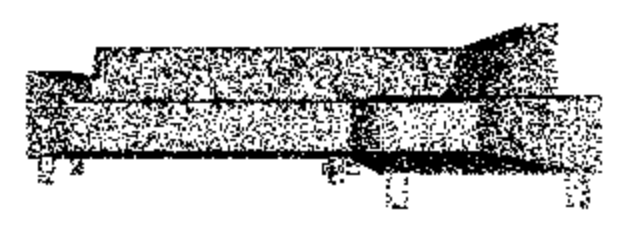}
    \hspace{1.6em}
    \includegraphics[width=0.30\linewidth,height=0.15\linewidth]{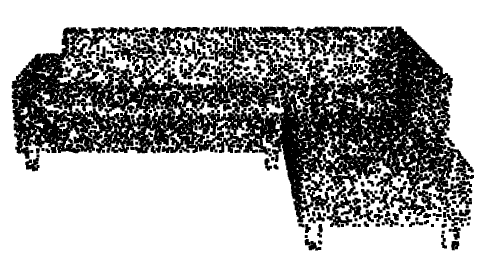}
  \end{minipage}
& 
  \begin{minipage}{\linewidth}
    \includegraphics[width=0.22\linewidth,height=0.15\linewidth]{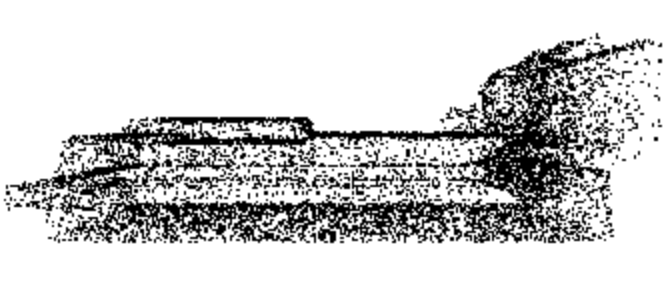}
    \hspace{1.6em}
    \includegraphics[width=0.25\linewidth,height=0.15\linewidth]{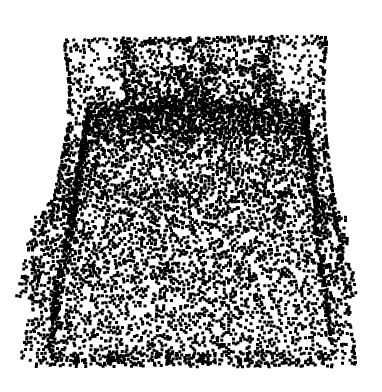}
  \end{minipage}
\\ \midrule
Prompt & What is this? & This is an object of  \\
Ground Truth & Sofa & Bed  \\ 
 \midrule
InstructBLIP-13B~\cite{dai2024instructblip}  & sofa & person cutting meat into small pieces with a large knife \\ 
LLaVA-13B\cite{liu2024visual}  & The image is a black and white drawing of a couch. & The image is a black and white depiction of a chair. \\ 
Point-Bind LLM~\cite{guo2023point}  & This is a drawing of a person laying on a couch, with a dog nearby. & This is an object of art. \\
PointLLM-13B~\cite{xu2023pointllm} & This 3D object is a comfortable, black, three-seater sofa. & This is a model of a neatly made bed, primarily composed of a solid black mattress. \\
\textbf{MiniGPT-3D} & \textbf{This is a 3D model of a {\color{LightGreen}modern, sleek black sofa}. The sofa is designed with a minimalist aesthetic, featuring clean lines and a neutral color palette. It is likely to be used in contemporary {\color{LightGreen}interior design, offering comfort and style.}} & \textbf{A 3D model of a {\color{LightGreen}bed} with a {\color{LightGreen}black mattress.} The bed is characterized by its {\color{LightGreen}rectangular shape} and the mattress, which is {\color{LightGreen}black in color.} The bed has {\color{LightGreen}a headboard and footboard}, both of which are also black. This model can be {\color{LightGreen}used in various digital platforms such as video games, animations, or virtual reality experiences.}} \\

\toprule

Samples 3, 4 & 
  \begin{minipage}{\linewidth}
    \includegraphics[width=0.25\linewidth,height=0.15\linewidth]{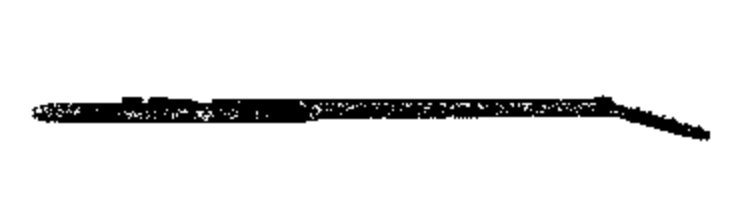}
    \hspace{1.6em}
    \includegraphics[width=0.18\linewidth,height=0.15\linewidth]{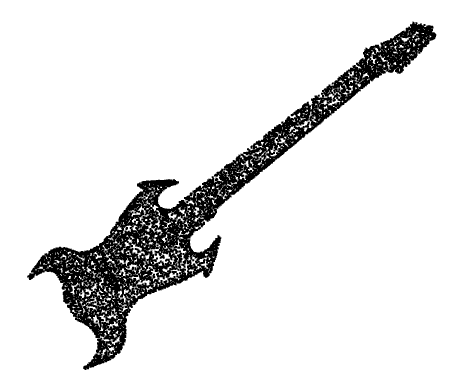}
  \end{minipage}
& 
  \begin{minipage}{\linewidth}
    \includegraphics[width=0.18\linewidth,height=0.15\linewidth]{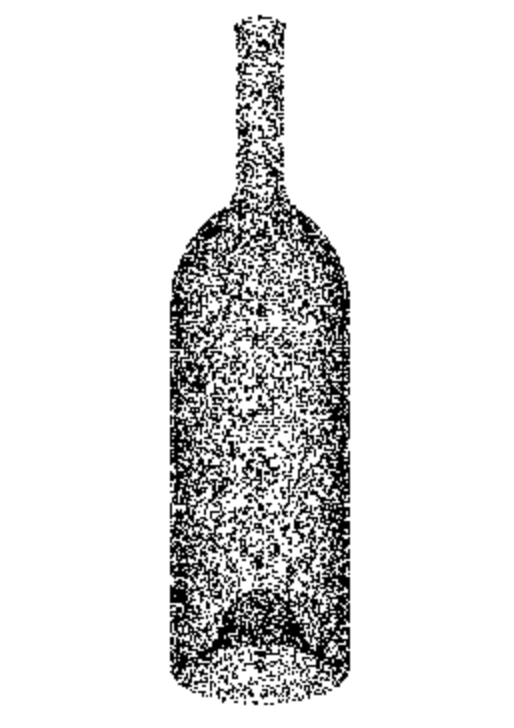}
    \hspace{1.6em}
    \includegraphics[width=0.18\linewidth,height=0.15\linewidth]{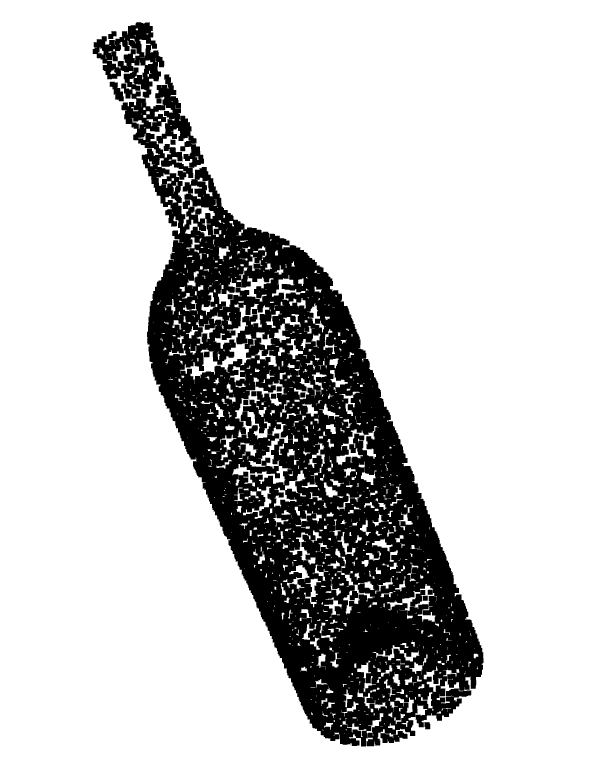}
  \end{minipage}
\\ \midrule
Prompt & What is this? & This is an object of  \\
Ground Truth & Guitar & Bottle \\  
 \midrule
InstructBLIP-13B~\cite{dai2024instructblip}  & telescope & wine bottle \\ 
LLaVA-13B\cite{liu2024visual}  & The image is a close-up view of a black ink pen sitting on a white surface. & This image features a black and white depiction of a bottle. \\ 
Point-Bind LLM~\cite{guo2023point}  & This is a bird flying in the sky. & This is an object of a person holding a bottle of wine. \\
PointLLM-13B~\cite{xu2023pointllm} & This 3D model represents a black electric guitar equipped with a distinctive headstock. & This is a 3D model of a bottle, possibly intended for liquids. \\
\textbf{MiniGPT-3D} & \textbf{ This is a 3d model of a cartoon-style {\color{LightGreen}guitar}. the guitar is predominantly {\color{LightGreen}black in color}, with a distinctively {\color{LightGreen}large headstock and a long neck}. the body of the guitar is also black, and it has a unique design with a large, flat headstock and a long neck. the guitar is likely to be used in animated films, video games, or other digital media.} & \textbf{A 3D model of a {\color{LightGreen}bottle}, predominantly {\color{LightGreen}black in color}. The bottle is {\color{LightGreen}cylindrical in shape, with a narrow neck and a wide base}. The bottle appears to be made of {\color{LightGreen}plastic or glass}, which is a common material used in the production of bottles. The bottle could be used for various purposes such as {\color{LightGreen}storing liquids, serving beverages, or as a decorative item.}} \\

\toprule
Samples 5, 6 & 
  \begin{minipage}{\linewidth}
    \includegraphics[width=0.25\linewidth,height=0.15\linewidth]{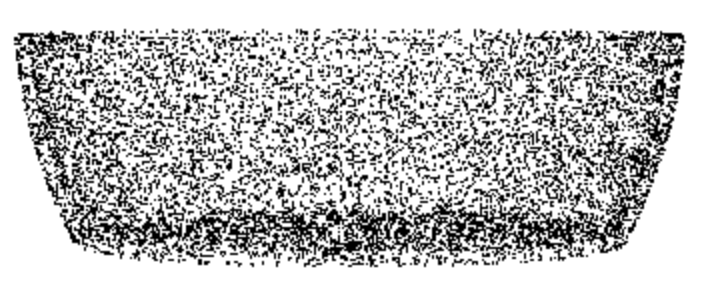}
    \hspace{1.6em}
    \includegraphics[width=0.18\linewidth,height=0.15\linewidth]{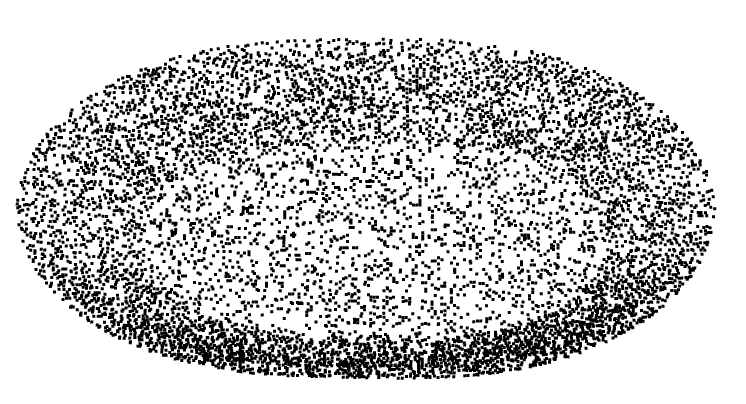}
  \end{minipage}
& 
  \begin{minipage}{\linewidth}
    \includegraphics[width=0.18\linewidth,height=0.15\linewidth]{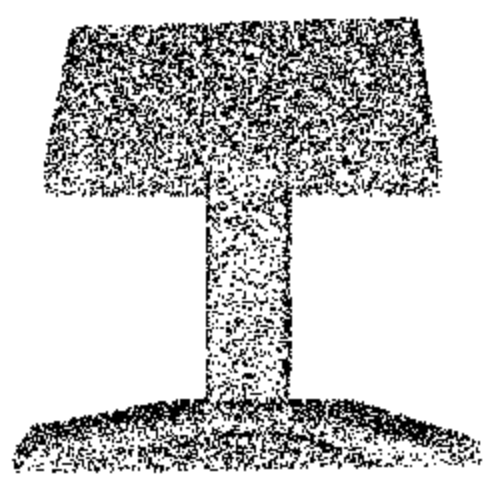}
    \hspace{1.6em}
        \includegraphics[width=0.18\linewidth,height=0.15\linewidth]{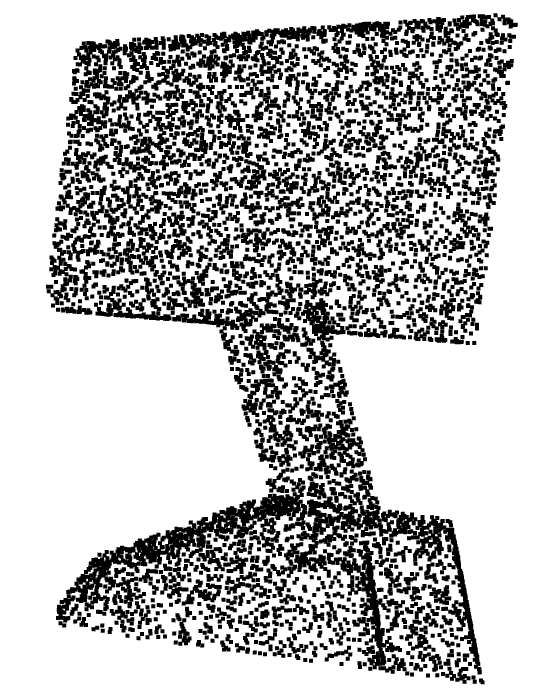}
  \end{minipage}
\\ \midrule
Prompt & What is this? & This is an object of  \\
Ground Truth & Bathtub & Monitor \\  
 \midrule
InstructBLIP-13B~\cite{dai2024instructblip}  & table & the foot \\ 
LLaVA-13B\cite{liu2024visual}  & The image features a black and white, close-up view of a grater, which is an object used for grating cheese or other food items. & The image features a small table with a black base and a white or light-colored top. \\ 
Point-Bind LLM~\cite{guo2023point}  & This is a bowl or a vase, which is oval-shaped and has a smooth surface. & This is an object of art, a sculpture, or a decorative piece. \\
PointLLM-13B~\cite{xu2023pointllm} & The 3D object model represents a realistic, dark grey bathtub. & This is a 3D model of a desktop computer monitor. \\
\textbf{MiniGPT-3D} & \textbf{This is a 3D model of a cartoon-style {\color{LightGreen}bathtub}. The bathtub is predominantly {\color{LightGreen}black in color} and has a unique{\color{LightGreen} design with a curved shape}. It is likely to be used in animated films, video games, or children's content.} & \textbf{A 3D model of {\color{LightGreen}a sleek, modern black computer monitor}. The monitor is {\color{LightGreen}rectangular in shape} and has a {\color{LightGreen}glossy finish}, giving it a sleek and professional look. It is equipped with a high-resolution display, suggesting it's suitable for tasks that require high-quality visuals, such as {\color{LightGreen}graphic design, video editing, or gaming}. The monitor is {\color{LightGreen} mounted on a sturdy stand, providing stability and convenience}.} \\
\bottomrule

\end{NiceTabular}
\end{table*}

\subsection{Training Details}
This section presents the training details of MiniGPT-3D, encompassing the training settings, model parameter, and the variation in loss across the four training stages.

\paragraph{\textbf{Training Settings}}

\begin{table*}[t]
  \centering

  \caption{Detailed training settings.}

  \resizebox*{\linewidth}{!}{
\begin{NiceTabular}{cc|cccc}
\toprule
\multicolumn{2}{c|}{\textbf{Setting}}  & \textbf{Stage I}  & \textbf{Stage II}  & \textbf{Stage III}  & \textbf{Stage  IV}  \\ \midrule
\multicolumn{2}{c|}{Dataset}  & \multicolumn{4}{c}{Point-text Instruction Dataset~\cite{xu2023pointllm}}  \\
\multicolumn{2}{c|}{Dataset Types}   & Brief Caption   & Brief Caption   & \begin{tabular}[c]{@{}c@{}}Detailed Caption\\ \& Conversation\end{tabular}   & \begin{tabular}[c]{@{}c@{}}Detailed Caption\\ \& Conversation\end{tabular}   \\
\multicolumn{2}{c|}{Dataset Scale}   & 660 k  & 660 k  & 70 k   & 70 k   \\ \midrule
  & Batch Size   & 9 & 9 & - & - \\
\multirow{-2}{*}{Brief Caption}  & Sample Ratio   & 1 & 1 & - & - \\ \cline{2-2}
  & Batch Size   & - & - & 6 & 6 \\
\multirow{-2}{*}{Detailed Caption} & Sample Ratio   & - & - & 2 & 2 \\ \cline{2-2}
  & Batch Size   & - & - & 10   & 10   \\
\multirow{-2}{*}{Single-round Conversation}   & Sample Ratio   & - & - & 3 & 3 \\ \cline{2-2}
  & Batch Size   & - & - & 4 & 4 \\
\multirow{-2}{*}{Multi-round Conversation}  & Sample Ratio   & - & - & 3 & 3 \\ \midrule
\multicolumn{2}{c|}{Max Epoch}   & 1 & 1 & 3 & 1 \\
\multicolumn{2}{c|}{Iterations Per Training Epoch}   & 70000  & 70000  & 10000  & 10000  \\
\multicolumn{2}{c|}{Learn Rate Scheduler}   & \multicolumn{4}{c}{linear\_warmup\_cosine\_lr}   \\
\multicolumn{2}{c|}{Initialized Learn Rate}   & 0.00003  & 0.00003  & 0.00001  & 0.000005 \\
\multicolumn{2}{c|}{Min Learn Rate}  & 0.00001  & 0.00001  & 0.000001 & 0.000001 \\
\multicolumn{2}{c|}{Warmup Learn Rate} & 0.000001 & 0.000001 & 0.000001 & 0.000001 \\
\multicolumn{2}{c|}{Warmup Steps}  & 7000   & 7000   & 3000   & 1000   \\
\multicolumn{2}{c|}{Weight decay}  & 0.05   & 0.05   & 0.05   & 0.05   \\ \midrule
  & Point Number   & 8192   & 8192   & 8192   & 8192   \\
  & Point Group Size & 32   & 32   & 32   & 32   \\
  & Point Patch  & 512  & 512  & 512  & 512  \\
  & Hidden Size  & 384  & 384  & 384  & 384  \\
  & Head of Attention   & 6 & 6 & 6 & 6 \\
\multirow{-6}{*}{Point Cloud Encoder} & Number of Layer   & 12   & 12   & 12   & 12   \\ \midrule
  & Number of Layer   & 2 & 2 & 2 & 2 \\
\multirow{-2}{*}{Point Cloud Projection Layer} &  Dimension & \begin{tabular}[c]{@{}c@{}}384-\textgreater{}768;\\ 768-\textgreater{}1408\end{tabular} & \begin{tabular}[c]{@{}c@{}}384-\textgreater{}768;\\ 768-\textgreater{}1408\end{tabular} & \begin{tabular}[c]{@{}c@{}}384-\textgreater{}768;\\ 768-\textgreater{}1408\end{tabular} & \begin{tabular}[c]{@{}c@{}}384-\textgreater{}768;\\ 768-\textgreater{}1408\end{tabular} \\ \midrule
  & Router Type & - & - & - & Sparse Router~\cite{shazeer2017outrageously}   \\
  & Top Experts  & - & - & - & 2 \\
  & Number of Query Experts & - & - & - & 8 \\
  & Number of Expert Router Layer  & - & - & - & 2 \\
\multirow{-5}{*}{Mixture of Query Experts}   & Dimension of Expert Router Layer & - & - & - & \begin{tabular}[c]{@{}c@{}}768-\textgreater{}256;\\ 256-\textgreater{}8\end{tabular}  \\ \midrule
  & Rank of LoRA   & - & 8 & 8 & 8 \\
  & Alpha of LoRA  & - & 16   & 16   & 16   \\
  & Number of Layer   & 12   & 12   & 12   & 12   \\
  & Head of Attention   & 12   & 12   & 12   & 12   \\
\multirow{-5}{*}{Q-Former} & Hidden Size  & 768  & 768  & 768  & 768  \\ \midrule
  & Number of Layer   & 2 & 2 & 2 & 2 \\
\multirow{-2}{*}{Modality Projector}  & Dimension & \begin{tabular}[c]{@{}c@{}}768-\textgreater{}4096;\\ 4096-\textgreater{}2560\end{tabular} & \begin{tabular}[c]{@{}c@{}}768-\textgreater{}4096;\\ 4096-\textgreater{}2560\end{tabular} & \begin{tabular}[c]{@{}c@{}}768-\textgreater{}4096;\\ 4096-\textgreater{}2560\end{tabular} & \begin{tabular}[c]{@{}c@{}}768-\textgreater{}4096;\\ 4096-\textgreater{}2560\end{tabular} \\ \midrule
  & Rank of LoRA   & 64   & 64   & 64   & 64   \\
  & Alpha of LoRA  & 16   & 16   & 16   & 16   \\
  & Number of Layer   & 32   & 32   & 32   & 32   \\
  & Head of Attention   & 32   & 32   & 32   & 32   \\
\multirow{-5}{*}{Large Lanuguage Model Backbone} & Hidden Size  & 2560   & 2560   & 2560   & 2560   \\ \bottomrule
\end{NiceTabular}
  }
  \label{sup_tab:all_setting}%
\end{table*}%

Table~\ref{sup_tab:all_setting} shows the detailed training settings for MiniGPT-3D.
Specifically, we use the point-text instruction dataset~\cite{xu2023pointllm} as the training dataset, encompassing 660k brief captions and 70k detailed captions \& conversations.
Within this setup, stages I and II employ the brief captions as their training dataset, while detailed captions \& conversations are utilized in stages III and IV.
Notably, stages III and IV utilize different types of training data from detailed captions \& conversations based on a specific sampling ratio.
For optimization, we adopt the AdamW optimizer with a weight decay of 0.05 and employ a cosine decay with a linear warm-up learning rate schedule. The initial learning rate gradually decreases as the training stage progresses.

Regarding the hyperparameters of model components, the point cloud encoder is configured consistently with Point-BERT~\cite{yu2022point}, receiving point cloud data inputs of 8192 points.
The point cloud projection layer consists of a two-layer MLP network that transforms the 384-dimensional features output from the point cloud encoder to the input dimension of 1408 for the Value and Key layers in Q-Former~\cite{li2023blip}.
Our proposed Mixture of Query Experts (MQE) comprises eight query experts and an expert router. The expert router includes a two-layer MLP network and a softmax operation, outputting the probability distribution for activating the eight query experts. We activate the two experts with the highest probabilities in our experiments.
Q-Former consists of 12 blocks, with each attention module containing 12 attention heads. LoRA~\cite{hu2021lora} is used for efficiently fine-tuning the  Q-Former, where the rank and alpha of LoRA are set to 8 and 16, respectively.
The modality projector consists of a two-layer MLP that transforms the 768-dimensional point cloud queries output from Q-Former to 2560-dimensional point tokens.
The large language model backbone comprises 32 blocks. We efficiently fine-tune the LLM using LoRA, with the rank and alpha of LoRA set to 64 and 16, respectively.

\begin{figure*}[t]
\centering
\begin{subfigure}[b]{0.45\linewidth}
    \centering
  \includegraphics[width=0.83\linewidth]{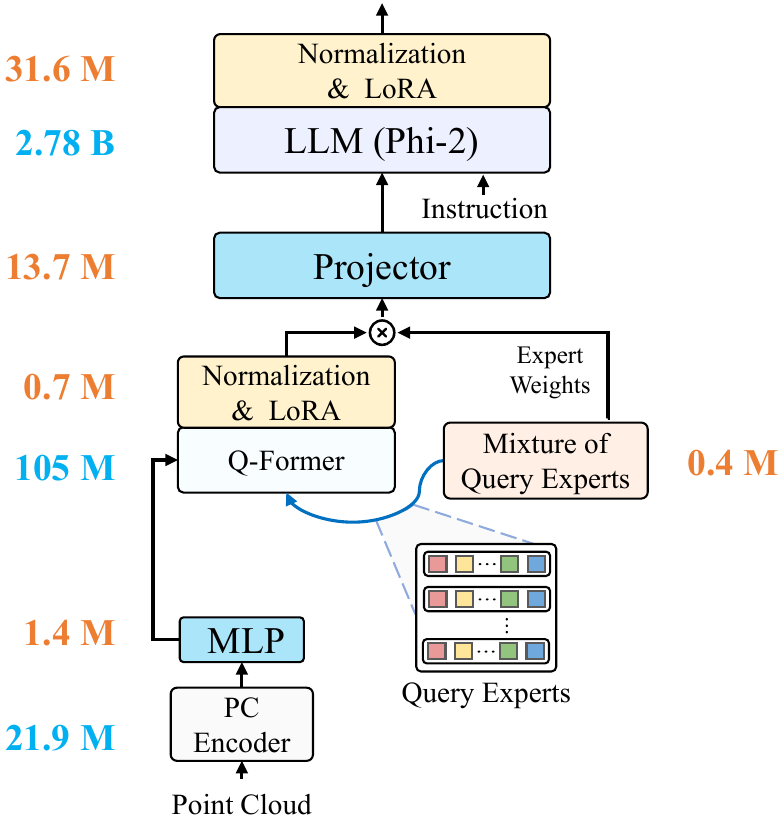}
  \caption{Architecture, module parameters of MiniGPT-3D.}
  \label{sup_fig: para}
\end{subfigure}
\hfill
\begin{subfigure}[b]{0.45\linewidth}

    \begin{tabular}{lrcr}
    \toprule
    \textbf{\begin{tabular}[c]{@{}l@{}}Trainable\\ Module\end{tabular}}  & \textbf{Params}   & \textbf{\begin{tabular}[c]{@{}c@{}}Frozen\\ Module\end{tabular}}  & \textbf{Params}   \\ \midrule
    \rowcolor[HTML]{EFEFEF} 
    \begin{tabular}[c]{@{}l@{}}Point Cloud Projection \\ Layer (MLP)\end{tabular} &  {\color[HTML]{ff8954} 1.4 M}  & \begin{tabular}[c]{@{}c@{}}PC \\  Encoder\end{tabular}  & {\color[HTML]{02b7db} 21.9 M}  \\
    \begin{tabular}[c]{@{}l@{}}Norm \& LoRA\\ of Q-Former\end{tabular}  &  {\color[HTML]{ff8954} 0.7 M}  & Q-Former  & {\color[HTML]{02b7db} 105 M} \\
    \rowcolor[HTML]{EFEFEF} 
    Modality Projector  &  {\color[HTML]{ff8954} 13.7 M}  & LLM (Phi-2)  &{\color[HTML]{02b7db}  2780 M}  \\ 
    Mixture of Query Experts  &  {\color[HTML]{ff8954} 0.4 M}  & - & - \\ 
    \rowcolor[HTML]{EFEFEF} 
    Norm \& LoRA  of LLM  &  {\color[HTML]{ff8954} 31.6 M}  & -   & -  \\ \midrule
    \textbf{Total   Parameters}  & {\color[HTML]{ff8954} \textbf{47.8 M}} & -   & {\color[HTML]{02b7db} \textbf{2907 M}}  \\ \bottomrule
    \end{tabular}
      \caption{Parameters and trainability of modules in MiniGPT-3D.}
      \label{sup_tab:total_para}
    
\end{subfigure}
      \caption{Architecture, module parameters, and module trainability of MiniGPT-3D.
   {\color[HTML]{02b7db}Blue} and {\color[HTML]{ff8954}orange} fonts indicate non-trainable and  trainable parameters, respectively.}
    
\vspace*{15mm}
\end{figure*}

\paragraph{\textbf{Model Parameter}}
MiniGPT-3D boasts a total of 2.95 B model parameters, yet we only train 47.8 M  parameters on a single RTX 3090 (24G) GPU, which took 27 hours.
The specific trainable and frozen model modules are detailed in Figure~\ref{sup_fig: para} and Table~\ref{sup_tab:total_para}.
%

\paragraph{\textbf{Training Loss}}

\begin{figure*}[t]
\centering
    \setlength{\abovecaptionskip}{0.1cm} 
  \includegraphics[width=0.75\linewidth]{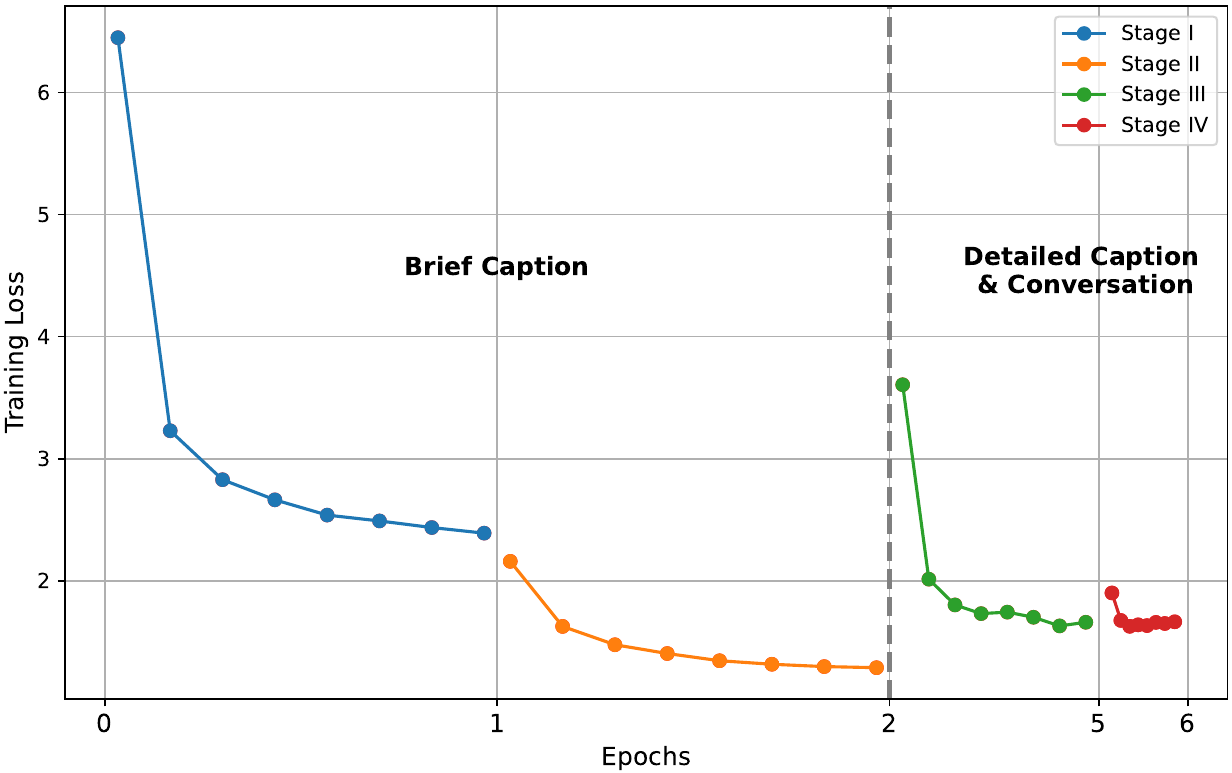}
  \caption{Changes in loss across the four training stages of MiniGPT-3D.}
  \label{sup_fig: all_stage_line}
\end{figure*}

Figure~\ref{sup_fig: all_stage_line} shows the changes in loss across the four training stages of MiniGPT-3D. The scale interval on the horizontal axis corresponds to the duration of training.
During stage I, though training  only point cloud projection layer (MLP), we observe a steady decrease in loss. 
During stage II, more modules are fine-tuned on the same dataset as stage I, enhancing the model's learning capacity, and leading to a continued decrease in loss from the end of stage I.
During stage III, the introduction of more challenging tasks temporarily increases the loss compared to the end of stage II, followed by a gradual reduction.
During stage IV, only MQE is trained. Since the expert router of MQE is trained from scratch, the loss suddenly increases compared to the end of stage III, but then gradually decreases to the same level or even lower.

\clearpage

\clearpage

\end{document}